\documentclass[10pt,twocolumn,letterpaper]{article}

\usepackage[pagenumbers]{cvpr} % To force page numbers, e.g. for an arXiv version

\usepackage{graphicx}
\usepackage{amsmath}
\usepackage{amssymb}
\usepackage{booktabs}

\usepackage{listings}
\usepackage{caption}
\usepackage{algorithm}
\usepackage{algpseudocode}
\usepackage{xcolor, cancel}
\usepackage{pifont}
\usepackage{soul}
\usepackage{pdflscape}

\makeatletter
\@namedef{ver@everyshi.sty}{}
\makeatother
\usepackage{tikz}

\usepackage{tabularx}
\usetikzlibrary{fit}

\newcounter{nodecount}
\newcommand\tabnode[1]{\addtocounter{nodecount}{1} \tikz \node  (\arabic{nodecount}) {#1};}

\tikzstyle{every picture}+=[remember picture,baseline]
\tikzstyle{every node}+=[anchor=base,minimum width=0.4cm,align=center,text depth=.25ex,outer sep=1.5pt]
\tikzstyle{every path}+=[thick, rounded corners]

\DeclareRobustCommand{\rchi}{{\mathpalette\irchi\relax}}
\newcommand{\irchi}[2]{\raisebox{\depth}{$#1\chi$}}

\definecolor{codegreen}{rgb}{0,0.5,0}
\definecolor{codegray}{rgb}{0.5,0.5,0.5}
\definecolor{codepurple}{rgb}{0.58,0,0.82}
\definecolor{backcolour}{rgb}{1.0,1.0,1.0}
\definecolor{link}{rgb}{1.0,1.0,1.0}
\lstdefinestyle{mystyle}{
    backgroundcolor=\color{backcolour},   
    commentstyle=\color{codegreen},
    keywordstyle=\color{magenta},
    numberstyle=\tiny\color{codegray},
    stringstyle=\color{codepurple},
    basicstyle=\ttfamily\footnotesize,
    breakatwhitespace=false,         
    breaklines=true,                 
    keepspaces=true,                 
    numbers=left,                    
    numbersep=5pt,                  
    showspaces=false,                
    showstringspaces=false,
    showtabs=false,                  
    tabsize=1,
    float=tp,
  floatplacement=tbp
}
\definecolor{black}{RGB}{87,100,120}
\definecolor{blue}{RGB}{33,186,207}

\definecolor{light_orange}{RGB}{252,219,191}
\definecolor{dark_orange}{RGB}{237,139,55}
\definecolor{cyan}{RGB}{123,250,241}

\DeclareCaptionFormat{mylst}{\hrule#1#2#3\hrule}
\newcommand{\cmark}{\ding{51}}%
\newcommand{\xmark}{\ding{55}}%
\newcommand*{\myalign}[2]{\multicolumn{1}{#1}{#2}}

\newcommand{\J}{\mathcal{J}}%
\newcommand{\JandF}{\mathcal{J}\&\mathcal{F}}%
\newcommand{\F}{\mathcal{F}}%

\definecolor{lightred}{rgb}{238,208,219}
\usepackage{rotating}
\usepackage{multirow}
\usepackage{tikz}
\usepackage{mathabx}

\usepackage{mathtools}
\DeclarePairedDelimiterX{\infdivx}[2]{(}{)}{%
  #1\;\delimsize\|\;#2%
}

\DeclarePairedDelimiter{\norm}{\lVert}{\rVert}

\newcommand{\Real}{{\rm I\!R}}

\newcommand{\Renyi}{{R\'enyi}}

\usepackage[pagebackref,breaklinks,colorlinks]{hyperref}

\usepackage{bookmark}

\usepackage[capitalize]{cleveref}
\crefname{section}{Sec.}{Secs.}
\Crefname{section}{Section}{Sections}
\Crefname{table}{Table}{Tables}
\crefname{table}{Tab.}{Tabs.}

\def\cvprPaperID{9632} % *** Enter the CVPR Paper ID here
\def\confName{CVPR}
\def\confYear{2023}

\begin{document}

\title{MobileVOS: Real-Time Video Object Segmentation\\
Contrastive Learning meets Knowledge Distillation}

\author{Roy Miles\thanks{Work done while being an intern at Samsung Research UK}
\hspace{6pt}
%\and
Mehmet Kerim Yucel
\hspace{6pt}
%\and
Bruno Manganelli
\hspace{6pt}
%\and
% \textsuperscript{1}
Albert Sa\`{a}-Garriga
\hspace{6pt}
\vspace{.6ex}
\\
% \textsuperscript{1} Samsung Research UK
Samsung Research UK
%\and
% \hspace{12pt}
% \textsuperscript{2}Imperial College London
}

\maketitle

%%%%%%%% ABSTRACT
\begin{abstract}
This paper tackles the problem of semi-supervised video object segmentation on resource-constrained devices, such as mobile phones. We formulate this problem as a distillation task, whereby we demonstrate that small space-time-memory networks with finite memory can achieve competitive results with state of the art, but at a fraction of the computational cost (32 milliseconds per frame on a Samsung Galaxy S22). Specifically, we provide a theoretically grounded framework that unifies knowledge distillation with supervised contrastive representation learning. These models are able to jointly benefit from both pixel-wise contrastive learning and distillation from a pre-trained teacher. We validate this loss by achieving competitive $\mathcal{J}\&\mathcal{F}$ to state of the art on both the standard DAVIS and YouTube benchmarks, despite running up to $\times5$ faster, and with $\times32$ fewer parameters.
\end{abstract}

%%%%%%%%% BODY TEXT
\begin{figure*}[t]
\centering
    \includegraphics[width=1.\textwidth]{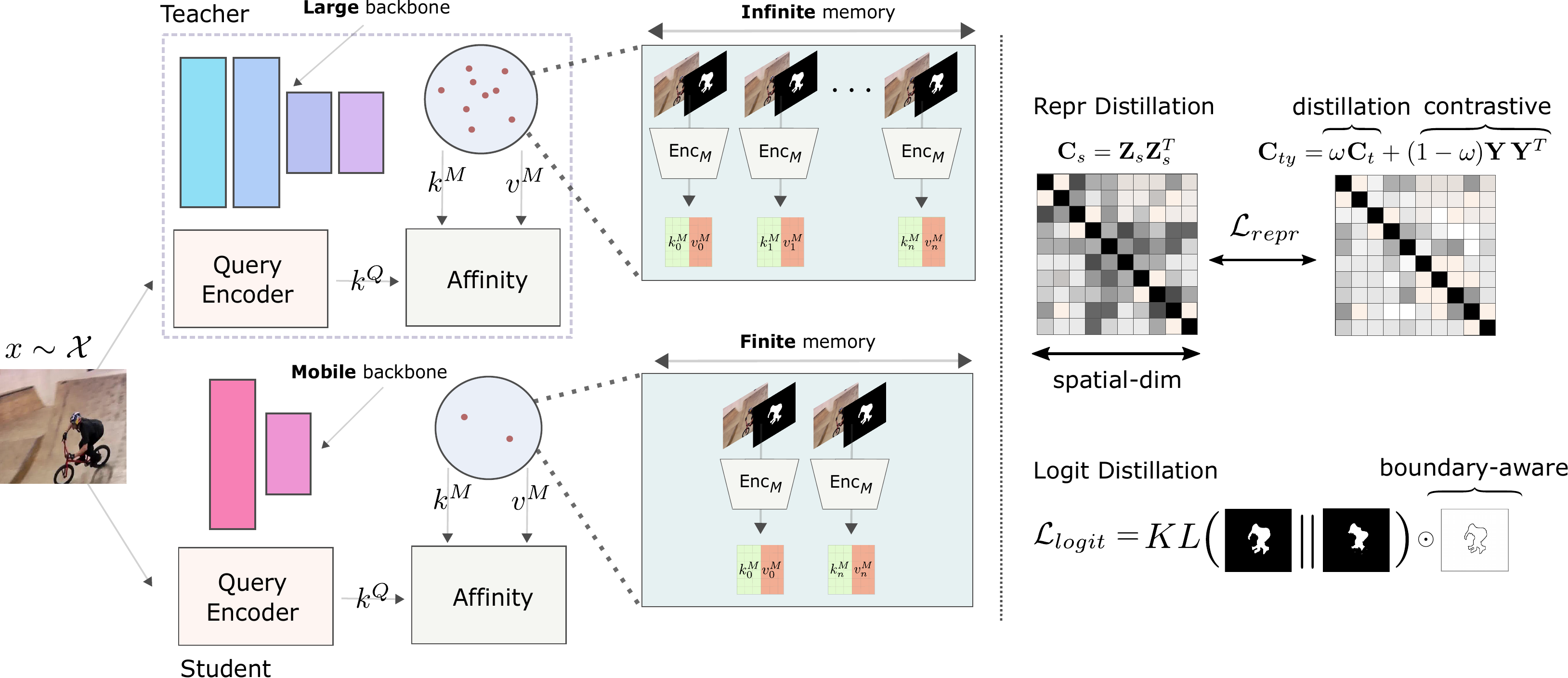}
    \caption{The proposed student-teacher framework using boundary-aware distillation. The teacher model utilises infinite memory, whereas the student model only retains memory from the previous and first frames/mask. Since the teacher has more memory, the task of distillation (equation \ref{eq:mi} and \ref{eq:logit}) is then to learn features which are consistent across multiple frames. The auxiliary objective of contrastive representation learning encourages discriminative pixel-wise features (equation \ref{eq:scl}) around the boundary of objects.}
    \label{fig:method}
\end{figure*}

\section{Introduction}
\label{sec:intro}
Video Object Segmentation (VOS) is a foundational task in computer vision, where the aim is to segment and track objects in a sequence of frames. VOS is the backbone of many applications such as video editing, autonomous driving, surveillance, and augmented reality \cite{wang2021survey}. In this work, we focus on semi-supervised VOS (SVOS), where the mask annotations for only the initial frame are provided. SVOS is a notoriously difficult task, where one needs to model the motion and appearance changes under severe occlusion and drifts, while performing well in a class-agnostic manner.

SVOS approaches can be divided into three main methods, namely, online fine-tuning~\cite{caelles2017oneOSVOS,hu2017maskrnn,li2018video,perazzi2017learning,voigtlaender2017online,mao2021joint}, object flow~\cite{cheng2017segflow,luiten2018premvos,oh2018fast,tsai2016video,xu2018dynamic,yang2018efficient}, and offline matching~\cite{feelvos,yang2020collaborativeCFBI,yang2021associating,chen2018blazinglyFast,hu2018videomatch}. Memory-based matching methods have shown impressive results in the last years. These approaches ~\cite{lu2020video,oh2019video,seong2020kernelized,cheng2021rethinking,wang2021swiftnet,xie2021efficient,liang2020video,liu2022global,cheng2022xmem,li2022recurrent} leverage memory banks to encode and memorize previous frames. They find feature correspondences between the memory and the current frame to predict the next object masks. Despite their good results, memory-networks tend to suffer from a trade-off between the memory usage and accuracy\cite{li2022recurrent}. Storing a reduced number of frames \cite{li2022recurrent,cheng2022xmem,wang2021swiftnet}, storing local segmentation features \cite{xie2021efficient,miao2022region} (i.e. features of only a part of the image), or segmenting only on keyframes \cite{xu2022accelerating}, have been proposed as methods to tackle this tradeoff. Recent methods are now accurate and fast on high-end GPUs, but there is no method in the literature that is capable of retaining state-of-the-art results while performing in real-time on mobile phones.
% red{but to the best of our knowledge}

Our work addresses real-time SVOS on resource-constrained devices. Specifically, we focus on memory-based networks and begin to bridge the gap between infinite memory and finite memory networks. In contrast with existing methods that do so by architectural \cite{li2022recurrent} and memory-bank design changes \cite{wang2021swiftnet,cheng2022xmem}, we adopt a novel approach that addresses the problem through knowledge distillation. Specifically, we propose a pixel-wise representation distillation loss that aims to transfer the structural information of a large, infinite memory teacher to that of a smaller, finite memory student. We then show that using a simple boundary-aware sampling strategy can improve training convergence. Finally, we provide a natural generalisation to encompass a supervised pixel-wise contrastive representation objective. Using ground-truth labels as another venue for structural information, we interpolate between representation distillation and contrastive learning with a hyper-parameter, and use it as a unified loss.  Our results on popular SVOS datasets show that we are competitive with state-of-the-art \cite{li2022recurrent}, while running $\times5$ faster and having $\times32$ fewer parameters. Our model, without any pruning or quantisation, runs comfortably in real-time on a mobile device. The summary of our main contributions is given as follows:
\vspace{-0.5em}

\begin{itemize}
\item We unify knowledge distillation and supervised contrastive learning to design a novel representation-based loss that bridges the gap between the large, infinite memory models and those with small, finite memory. Additionally, we show that a simple boundary-aware pixel sampling strategy can further improve the results and model convergence. \vspace{-0.2em}
\item Using this unified loss, we show that a common network design can achieve results competitive to the state-of-the-art,  while running up to $\times5$ faster and having $\times32$ fewer parameters. \vspace{-0.2em}
\item Without complex architectural or memory design changes, our proposed loss can unlock real-time performance (30FPS+) on mobile devices (i.e. Samsung Galaxy S22) while retaining competitive performance with state-of-the-art.  \end{itemize}

\section{Related Work}
\label{sec:related_work}
\noindent \textbf{Semi-Supervised Video Object Segmentation} Some SVOS methods perform offline pretraining and then online per-mask finetuning to learn object-specific representations \cite{caelles2017oneOSVOS,hu2017maskrnn,li2018video,perazzi2017learning,voigtlaender2017online,mao2021joint}. Such methods are naturally unfit for deployment due to the finetuning, which is done for each object mask given in the first frame. Addressing the online adaptation requirement of such methods are flow-based approaches \cite{cheng2017segflow,luiten2018premvos,oh2018fast,tsai2016video,xu2018dynamic,yang2018efficient}, which model SVOS as a temporal mask propagation problem. These methods, however, lack long-term context and are known to suffer from errors accumulated due to occlusion and drift. Detection-based methods address \cite{huang2020fastTemporalAggregation,li2018video,luiten2018premvos,Vock2019FastClouds} error accumulation via detectors, however, their performance is bound by the detector's performance. Matching-based methods~\cite{feelvos,yang2020collaborativeCFBI,yang2021associating,chen2018blazinglyFast,hu2018videomatch} match the features of the current and previous frame(s), and currently dominate the field due to the excellent results of memory-networks \cite{oh2019video}.

Following STM \cite{oh2019video}, many variants of memory-network based methods have emerged \cite{oh2019video,lu2020video,seong2020kernelized,cheng2021rethinking,wang2021swiftnet,xie2021efficient,liang2020video,liu2022global,cheng2022xmem,li2022recurrent}. STCN reformulated memory-networks and decoupled the masks and images in memory to increase efficiency \cite{cheng2021rethinking}. Per-Clip VOS \cite{park2022per} proposed to process whole clips instead of individual frames. GSFM \cite{liu2022global} proposed to leverage frequency-spectra to better model intra-frame spatial dependencies. XMem \cite{cheng2022xmem}  proposed a three layered memory bank scheme to perform equally well on long videos. QDMN \cite{liu2022learning} proposed a quality assessment module to select informative frames to store in the memory. Several methods proposed to leverage spatial redundancy to perform localized segmentation/matching \cite{xie2021efficient,miao2022region}. %Data augmentation regimes aiming to achieve robust matching have found use as well \cite{li2022recurrent,cho2022tackling}. 
RDE-VOS \cite{li2022recurrent} maintains a single template in memory whereas SwiftNet \cite{wang2021swiftnet} performs pixel-wise updates to ensure a constant memory cost. Despite the advances in efficiency, existing methods are far from feasible for mobile deployment, either due to subpar accuracy or excessive resource consumption.

\noindent \textbf{Knowledge Distillation} (KD) uses the predictions of a larger model (teacher) to provide additional supervision to a much smaller model (student) during training. KD is a common technique for model compression, making it useful for deployment scenarios. Its usage was first proposed in the context of image classification~\cite{Hinton2015DistillingNetwork} and has since been extended to depth estimation\cite{yucel2021real,liu2020structured}, segmentation\cite{Chen2021DistillingReview}, object detection \cite{shu2021channel} and image translation\cite{ZhangWaveletDistillation} tasks. In addition to simple logit distillation \cite{Hinton2015DistillingNetwork}, some of the most successful frameworks look at the representations before the final fully connected layer since they can preserve some of the structural information about the input. CRD~\cite{Tian2019ContrastiveDistillation} extended the supervised contrastive losses to incorporate the teachers' representations. This was later extended by both WCoRD~\cite{Chen2020WassersteinDistillation} and ITRD~\cite{Miles2021InformationDistillationBMVC} using the Wasserstein distance and an information theoretic framework, respectively. Although there is a rich literature on distillation for dense prediction tasks \cite{Chen2021DistillingReview,ZhangWaveletDistillation,shu2021channel,liu2020structured}, our work, to the best of our knowledge, is the first to leverage KD for SVOS. 

\noindent \textbf{Self-Supervision} (SS) aims to learn a stable representation from unlabelled data by leveraging known invariances through the use of data augmentations. They can be grouped into contrastive\cite{Chen2020ARepresentations} and de-correlation based \cite{Zbontar2021BarlowReduction, Bardes2021VICReg:Learning} methods. However, recent work has demonstrated the duality between these two paradigms under mild assumptions\cite{GarridoDualityContrastive}. On a similar note, knowledge distillation shares very close similarities to self-supervision\cite{Xu2020KnowledgeSelf-supervision} and its usage in SVOS has motivated this work. Although several works have focused on self-supervision for video object segmentation \cite{zhu2020self,yang2021self,zhu2021self}, unsupervised distillation for segmentation~\cite{Hamilton2022UnsupervisedCorrespondences}, or general self-distillation~\cite{Furlanello2018BornNetworks,Miles2020CascadedSelf-distillation}, there is no work exclusively focusing on self-supervised contrastive learning for SVOS. In addressing this gap we also provide a very general framework that is unified with knowledge distillation for improving the efficacy of training small mobile models. Theoretically, we also provide a different perspective of our loss and its direct equivalence to a generalised notion of mutual information.

Our work is the first to achieve real-time performance on mobile devices and the first to introduce distillation in the context of SVOS. RDE-VOS \cite{li2022recurrent} introduces a guidance loss, which can be seen as a form of memory distillation, but they also impose significant architectural changes. On the other hand, MobileVOS  performs competitively while running up to $\times5$ faster with $\times32$ fewer parameters.
\vspace{-0.16em}
%
%
%-------------------------------------------------------------------------
%
\vspace{-0.8em}
\section{Method}
\label{sec:method}
In this work we begin to bridge the performance gap between small finite memory space-time networks and their large infinite memory counterparts. This work is the first to open the possibility for real-time high-quality video object segmentation on resource-constrained devices, such as mobile phones. We propose to approach this problem through a knowledge distillation perspective and in doing so we develop a novel unification with contrastive learning. In the following sections, we introduce the distillation loss and its connection with information theory. Subsequent sections then show a series of natural extensions to this loss through boundary-aware sampling and a joint contrastive objective. We note that our framework is also very general: it can be applied to any other SVOS method and is likely applicable to other dense prediction tasks.

\subsection{Representation Distillation}
In light of recent works in the field of knowledge distillation~\cite{Tian2019ContrastiveDistillation, Miles2021InformationDistillationBMVC, Chen2020WassersteinDistillation}, we propose to use the representation right before the final fully-connected layer. Using earlier layers may hurt the student's performance\cite{Tian2019ContrastiveDistillation} through differing inductive biases, whereas the collapsed output space will have lost a lot of the structural information that can benefit the knowledge transfer.

More concretely, we propose a pixel-wise representation distillation loss that can transfer the structural information between these two models. The same augmented frames are fed into both networks and a distillation loss is applied at the representation level (see Figure \ref{fig:method}). To extract structural information, we propose to construct correlation matrices, $\mathbf{C}_s,\mathbf{C}_t \in \Real^{HW \times HW}$, from the two sets of representations, where $HW$ is the spatial size. These matrices will then capture the relationship between all pairs of pixels.
\vspace{-0.5em}

\begin{align}
    \mathbf{C}_{s} = \mathbf{Z}_{S}\mathbf{Z}_{S}^T, \;\;\; \mathbf{C}_{t} = \mathbf{Z}_{T}\mathbf{Z}_{T}^T
\end{align}

where $\mathbf{Z}_S \in \Real^{HW \times d_s}$ and $\mathbf{Z}_T \in \Real^{HW \times d_t}$ denote the $d$-dimensional $L2$ normalised student and teacher representations provided before the final point-wise convolution and upsampling layers. Motivated by recent improvement in the choice of distance metrics for distillation tasks~\cite{Miles2021InformationDistillationBMVC}, we derive the final representation loss as follows:

\vspace{-0.5em}
\begin{align}
    \mathcal{L}_{repr} &= \frac{1}{\mid \mathbf{C}_s \mid} \bigg( \log_2 \norm{\mathbf{C}_s}^2 - \log_2 \norm{\mathbf{C}_{s} \odot \mathbf{C}_{t}}^2 \bigg)
    \label{eqn:loss}
\end{align}

where $\odot$ is the Hadamard product and $\norm{\cdot}$ is the Frobenius norm. The two components of this loss can be interpreted as a regularisation term, and a correlation alignment between the two models, while the $\log_2$ operator is used here to improve the robustness to spurious correlations. An interesting property of this loss formulation is that it is equivalent to maximising the pixel-wise mutual information between the student and teacher representations (see Supplementary for derivation).

\vspace{-0.5em}
\begin{align}
    \mathcal{L}_{Distill} &= \mathbf{H}_{2}(\mathbf{Z}_S) - \mathbf{H}_{2}(\mathbf{Z}_S ; \mathbf{Z}_T) \\
    &= -\mathbf{I}_{2}(\mathbf{Z}_S ; \mathbf{Z}_T) \label{eq:mi}
\end{align}

where $\mathbf{H}_2$ and $\mathbf{I}_2$ are matrix-based estimators~\cite{SanchezGiraldo2013InformationKernels} resembling \Renyi's entropy and mutual information of order 2 respectively. From this perspective, the two terms can instead be interpreted as maximising the joint entropy subject to an entropy regularisation.

\subsection{Unification with contrastive learning}
The effectiveness of knowledge distillation alone can be dependant on the relative capacity gap between the student and teacher models. More formally, this training regime scales poorly as the capacity gap diminishes. To address this constraint, we propose a vast generalisation of equation \ref{eqn:loss} to encompass pixel-wise contrastive learning. Other works~\cite{Xu2020KnowledgeSelf-supervision} have already shown that using self-supervision as an auxiliary task can improve conventional distillation, but there has yet to be a single unification of the two.

Since we have the dense ground truth classes, we can construct an additional target correlation matrix $\left(\mathbf{C}_y\right)_{ij} \in \{0, 1\}$ as follows:
\begin{align}
    \mathbf{C}_y = \mathbf{Y}\mathbf{Y}^T
\end{align}

where $\mathbf{Y} \in \Real^{HW \times 2}$ are the spatially-downsampled, one-hot-encoded labels for the 2 classes (object and background). We can then couple the two target correlation matrices to provide a way of interpolating between these two training regimes.

\vspace{-0.5em}
\begin{align} \label{eq:omega}
    \mathbf{C}_{ty} &= 	\omega\mathbf{C}_{t} + (1 - \omega)\mathbf{Y}\mathbf{Y}^T
\end{align}

where $	\omega \in [0, 1]$ is a hyperparameter and $\mathbf{C}_{ty}$ can now be substituted into equation~\ref{eqn:loss}. % In the case where $\omega = 0$, we have a purely contrastive objective, and the case where $\omega = 1$, we are only performing distillation.
By considering a representation $\mathbf{Z}$ and the case where $\omega = 0$, the loss will now reduce to a familiar supervised contrastive learning (SCL) setting~\cite{Khosla2020SupervisedLearning}.

\begin{align}
    \mathcal{L}_{SupCon} = -\frac{1}{\mid \mathbf{C}_s \mid} \log_2 \sum_i \frac{\sum_{j \in \mathcal{P}_i} \textit{sim}(\mathbf{Z}_{i}, \mathbf{Z}_{j})}{\sum_k \textit{sim}(\mathbf{Z}_{i}, \mathbf{Z}_{k})} \label{eq:scl}
\end{align}

where $\textit{sim}$ is the cosine similarity between two individual pixels in a representation and $\mathcal{P}_i$ is the set of positive indices for $i$-th pixel (see supplementary). Intuitively, for a given pixel, the numerator attracts the positives, while the denominator repels the negatives. The connection between these two regimes is illustrated in the following diagram:

\vspace{-0.5em}
\begin{align}
    \mathcal{L}_{SupCon} \xleftarrow[]{\omega = 0} \mathcal{L}_{repr} \xrightarrow[]{\omega = 1} \mathcal{L}_{Distill}
\end{align}

where the choice of $\omega$ is motivated by the availability and relative performance of a pre-trained teacher model. 

% # Region-masking
% boundary_mask = sobel(y) > threshold
% z_s = z_s[boundary_mask]
% z_t = z_t[boundary_mask]
% y_s = y_s[boundary_mask]
% y_t = y_t[boundary_mask]
% # $\color{codegreen}\omega$: Interpolation factor
% # $\color{codegreen}\tau$: Temperature
% manually transforms Algorithm 1.1 to 1

% # sample_boundary: Select edge pixels from near and on the boundary of the G.T. mask
%   # Boundary sampling conditioned on the G.T. mask
%   z_s = sample_boundary(z_s)
%   z_t = sample_boundary(z_t)
%   y = sample_boundary(y)
\renewcommand\thelstlisting{1}
\begin{figure}[t]
\centering
\begin{minipage}[t]{0.9\linewidth}
\begin{lstlisting}[basicstyle=\scriptsize\ttfamily, mathescape, language=Python, caption={PyTorch-style pseudocode for MobileVOS}, captionpos=t,
numberbychapter=false,label={alg:training_pseudo_code}]
# f_s, f_t: Student and teacher network
# y: Ground-truth one-hot encoded labels
# y_s, y_t: Student and teacher logits
# z_s, z_t: Student and teacher representations
for x, y in loader:
  # Forward pass
  z_s, y_s = f_s(x)
  z_t, y_t = f_t(x)
  z_s = embed_s(z_s)
  
  # Cross entropy loss
  loss = bootstrap_poly_cross_entropy(y_s, y)
  
  # Normalise representations
  z_s_norm = F.normalize(z_s, dim=1)
  z_t_norm = F.normalize(z_t, dim=1)
  
  # Compute correlation-matrices
  c_ss = matmul(z_s_norm, z_s_norm.T)
  c_tt = matmul(z_t_norm, z_t_norm.T)
  
  # Interpolate between KD and SCL
  y_d = downsample(y)
  yy = matmul(y_d, y_d.T)
  
  r = $\omega$ * c_tt + (1 - $\omega$) * yy
  loss += log2(c_ss.pow(2).sum()) / len(z_s) 
  loss -= log2((c_ss * r).pow(2).sum()) / len(z_s)
  
  # Logit distillation
  prob_s = softmax(y_s / $\tau$, dim=1)
  prob_t = softmax(y_t / $\tau$, dim=1)
  loss += kl(prob_t, prob_s)
  
  # Optimisation step
  loss.backward()
  optimizer.step()
\end{lstlisting}
\end{minipage}
\end{figure}
% \roy{Fix dimensions in pseudo-code as there is some implicit reshaping going on.}

\subsection{Boundary-aware sampling}
Most prediction errors occur on the boundary of objects (see Figure~\ref{fig:pixel_sampling}). Additionally, constructing the correlation matrix for all pixels is far too computationally expensive. Motivated by both of these two points, we propose to only sample pixels around and near the boundary of the objects. This sampling strategy restricts the distillation gradients to only flow through pixels that lead to downstream prediction errors, while also enabling a much more computationally efficient formulation. Since each frame also has a different boundary, the normalisation term (Equation \ref{eqn:loss}) will now average over the size of these object boundaries, thus allowing the loss to naturally uniformly weight both small and large objects evenly. We additionally observe that this modification can improve the overall model convergence, as shown in Figure \ref{fig:pixel_sampling}.

\subsection{Logit distillation}
An additional KL divergence term is introduced at the logit space between the two models, which is common in the distillation literature~\cite{Tian2019ContrastiveDistillation}.

\begin{align}
    % \mathcal{L}_{logit} &= \frac{1}{HW}\sum h(\mathbf{Y}) \cdot KL\left( \mathbf{p}_S(\tau) \; \lVert \; \mathbf{p}_T(\tau) \right)
    \mathcal{L}_{logit} &= \frac{1}{HW}\sum KL\left( \mathbf{p}_S(\tau) \; \lVert \; \mathbf{p}_T(\tau) \right) \label{eq:logit}
\end{align}

where $\mathbf{p}_S, \mathbf{p}_T$ are the student and teacher probabilities parameterised by a temperature term $\tau$ for softening ($\tau > 1$) or sharpening ($\tau < 1$) the two predictions. The summation indices have been omitted for brevity. The only distinction from conventional logit distillation is that we only use the pixels around or near the boundary of the objects. The final loss is given as follows:

\begin{align}
    \mathcal{L} = \mathcal{L}_{cross-entropy} + \mathcal{L}_{logit} + \mathcal{L}_{repr}
\end{align}

The complete training pipeline can be seen in Figure \ref{fig:method} and the PyTorch-style pseudo-code is given in Algorithm \ref{alg:training_pseudo_code}. For brevity, we omit the boundary-aware sampling in pseudo-code, but this can be straightforwardly implemented using a Sobel edge detector on the ground truth masks. 

\begin{figure}
\centering
\includegraphics[width=0.7\linewidth]{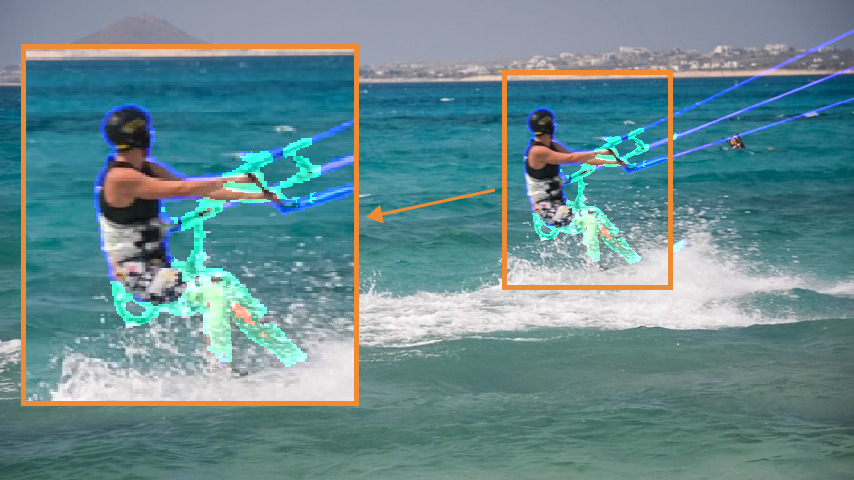}
\label{fig:region_errors}
\caption{Prediction errors, shown in \textcolor{cyan}{cyan}, can typically occur on the boundaries of the segmented object, thus motivating the emphasis on distilling and contrasting boundary pixels.}
\end{figure}

%-------------------------------------------------------------------------
% \vspace{-1em}
\section{Experiments}
\label{sec:experiments}
We evaluate MobileVOS on two standard benchmarks, namely DAVIS~\cite{davis16,davis17} and YouTube~\cite{youtubevos}. For all experiments, we train the student with an additional linear embedding layer and with a pre-trained STCN~\cite{cheng2021rethinking} as the teacher. Since most SVOS methods in the literature report the FPS metrics on different hardware, we further provide an additional fair comparison of our distilled models on the same sets of hardware. In these experiments, we consider both a server-grade GPU and a desktop-grade GPU, whereby our models are consistently at least twice as fast as competing methods for both long and short videos while being up to $\times32$ smaller. The smallest distilled model is then deployed on a mobile phone for real-time segmentation, which opens up the possibility for many new on-device applications.  Finally, we propose the use of model soups \cite{wortsman2022model}, for which are able to match the best performing RDE-VOS model that is specifically trained for YouTube.

\subsection{Datasets and Metrics}

\paragraph{DAVIS}
The DAVIS datasets are high-quality and high-resolution, densely-annotated videos for video object segmentation. DAVIS 2016~\cite{davis16} provides single-object segmentation, with a validation set of 20 videos, while DAVIS 2017~\cite{davis17} considers multi-object segmentation with both a validation and test set of 30 videos.

\vspace{-1em}
\paragraph{YouTube-VOS}
YouTube-VOS 2019~\cite{youtubevos} is a large-scale benchmark dataset for multi-object segmentation. It consists of 3,471 videos for training and 507 videos for validation, across 65 categories. There are also 26 additional unseen categories in the validation set.

\subsection{Implementation Details}
To demonstrate the simplicity of our proposed representation loss, we provide its pseudo-code in algorithm~\ref{alg:training_pseudo_code}. All of our experiments were trained using $4$ NVIDIA A10 GPUs with a batch size of 8. For evaluating the FPS metrics on the DAVIS datasets, we used a single NVIDIA A10 GPU with a batch-size of 1. Adam\cite{kingma2015adam} was used as the optimizer, with a learning rate of $1e-5$ and a weight decay of $1e-7$. The training details and architectures are primarily unchanged from that in the original STCN~\cite{cheng2021rethinking}, but are discussed in detail in the following sections.

\vspace{-1em}
\paragraph{Architecture modifications}
Motivated by the goal to deploy these models on mobile devices, we adopt a few architectural modifications to the original STCN architecture. The value encoder is replaced with a MobileNetV2\cite{Fox2018MobileNetV2:Bottlenecks} backbone, while the query encoder is either replaced with ResNet18\cite{He2015ResNetRecognition} or another MobileNetV2. The frame is also removed as an input to the value encoder and, unless otherwise specified, an ASPP module - commonly employed in the SVOS literature~\cite{li2022recurrent, wang2021swiftnet} - is used before the decoder to introduce more global context. Finally, we select a fixed queue length of 2, since it enabled real-time performance on a mobile device, while maintaining the top-end accuracy (see figure \ref{fig:mobile_perf}).

% \red{Finally, we settle on a fixed memory queue length of 2 as the sweetspot between inference speed and $\JandF$ metrics.}

\vspace{-1em}
\paragraph{Training details}
We follow the same training methodology proposed in STCN~\cite{cheng2021rethinking} except with two distinct modifications. Firstly, we extend the sequence length of videos to 5 to enable the student to learn features that are consistent across multiple frame predictions and secondly we use the poly cross-entropy loss~\cite{leng2022polyloss} with $\epsilon = 1$. The poly loss can be seen as a natural generalisation of the cross-entropy and focal loss~\cite{focal_loss}, which encourages the model to focus on the hard misclassified pixels.

\vspace{-1em}
\paragraph{Training stages}
All models are first trained on static images with synthetic deformations. The main training stage uses both the YouTube-VOS and DAVIS 2017 datasets for 600k iterations and with a batch size of 8. BatchNorm layers are also frozen throughout all training stages.

\subsection{Comparisons to State-of-the-art}
Similar to RDE-VOS~\cite{li2022recurrent}, we use \textbf{Constant Cost (CC)} to denote methods with finite memory during inference. We consider two different key encoder backbones, namely a ResNet18 and a MobileNetV2. For the MobileNetV2 model, we also show results with and without ASPP. % Unlike RDE-VOS~\cite{li2022recurrent}, we report the performance of the same model checkpoints across both the DAVIS and YouTube datasets. 
Since the ResNet architectures achieve close to the original STCN performance, we choose to use a purely contrastive loss i.e. $\omega = 0.0$. This helps avoid overfitting to the teacher's predictions on known classes and improve its generalisation to the unknown classes on YouTube. In contrast, for the much smaller MobileNet architectures, we use $\omega = 0.95$. These much smaller models are unlikely to overfit, and thus can benefit more from the distillation objective. It is worth noting that, in all cases, the models are also jointly trained with logit distillation, where $\tau = 0.1$.

\vspace{-0.5em}
\paragraph{DAVIS}
We compare MobileVOS against previous state-of-the-art methods on the DAVIS 2016 and DAVIS 2017 validation splits. The results for DAVIS 2016 are shown in Table \ref{table:davis16}, in which our best performing model is just $0.3\;\JandF$ shy of STCN and $0.2\;\JandF$ shy of RDE-VOS, while running $4 \times$ and $3 \times$ faster than these models, respectively.

Table \ref{table:davis17} shows our results on DAVIS 2017. We are highly competitive, where we are only slightly worse ($0.3\;\JandF$) than our STCN teacher model. Note that we are nearly $5 \times$ and $4 \times$ faster than STCN and RDE-VOS, which shows our runtime performance is still competitive despite tracking multiple objects. %\red{Update with new results?}

\begin{table}[!htbp]
    \centering
    \begin{tabular}{lccccc}
    \toprule
    \textbf{Method} & CC & $\JandF$ & $\J$ & $\F$ & FPS \\
    \midrule
    RMNet$^\dagger$~\cite{xie2021efficient} & \xmark & 88.8 & 88.9 & 88.7 & 11.9 \\
	STM$^\dagger$~\cite{oh2019video} & \xmark & 89.3 & 88.7 & 89.9 & 6.3 \\
% 	KMN$^\dagger$~\cite{seong2020kernelizedMemory} & \xmark & 90.5 & 89.5 & 91.5 & 8.4 \\
% 	LCM$^\dagger$~\cite{hu2021learning} & \xmark & 90.7 & 89.9 & 91.4 & 8.5 \\
% 	HMMN$^\dagger$~\cite{seong2021hierarchical} & \xmark & 90.8 & 89.6 & 92.0 & 10.0 \\
	MiVOS$^{\dagger*}$~\cite{cheng2021mivos} & \xmark & 91.0 & 89.7 & 92.4 & 16.9 \\
	STCN$^{\dagger*}$~\cite{cheng2021rethinking} & \xmark & 91.7 & \underline{90.4} & 93.0 & \underline{26.9} \\
	BATMAN~\cite{yu2022batman} & \xmark & \textbf{92.5} & \textbf{90.7} & \textbf{94.2} & - \\
	GSFM~\cite{liu2022global} & \xmark & 91.4 & 90.1 & 92.7 & $\approx$ 8.9 \\
	XMem$^{\dagger}$~\cite{cheng2022xmem} & \xmark & 91.5 & \underline{90.4} & 92.7 & \textbf{29.6} \\
	XMem$^{\dagger*}$~\cite{cheng2022xmem} & \xmark & 92.0 & \textbf{90.7} & \underline{93.2} & \textbf{29.6} \\
	\midrule
	GCNet~\cite{li2020fastGlobalContext} & \cmark & 86.6 & 87.6 & 85.7 & 25.0 \\
	CFBI$^\dagger$~\cite{yang2020collaborativeCFBI} & \cmark & 89.9 & 88.7 & 91.1 & 5.9 \\
	SwiftNet$^\dagger$~\cite{wang2021swiftnet} & \cmark & 90.4 & \textbf{90.5} & 90.3 & 25.0 \\
	RDE-VOS$^\dagger$~\cite{li2022recurrent} & \cmark & 91.1 & 89.7 & 92.5 & 35.0 \\
	RDE-VOS$^{\dagger*}$~\cite{li2022recurrent} & \cmark & \textbf{91.6} & 90.0 & \textbf{93.2} & 35.0 \\
% 	\midrule
	%\begin{tabular}{@{}c@{}}MobileVOS \\ \footnotesize ResNet18  \end{tabular} & \cmark & 90.7 & - & - & 108.8 \\
	%\footnotesize $\alpha = 0.95$ & \cmark & 90.7 & 89.8 & 91.7 & - \\
    & & & & & \\[-2ex]
	\myalign{l}{\hspace{-0.6em}\tabnode{MobileVOS}} & & & & & \\	
	\myalign{l}{\;\;\;\footnotesize ResNet18$^\dagger$} & \cmark & 90.6 & 89.7 & 91.6 & \textbf{100.1} \\

	\myalign{l}{\;\;\;\footnotesize ResNet18$^{\dagger*}$} & \cmark & \underline{91.4} & \underline{90.3} & \underline{92.6} & \textbf{100.1} \\
	
	\myalign{l}{\;\;\;\footnotesize $\rotatebox[origin=c]{180}{$\Lsh$}$ model soup$^\dagger$} & \cmark & 91.3 & 90.2 & 92.5 & \textbf{100.1} \\
	
    \myalign{l}{\;\;\;\footnotesize MobileNetV2$^\dagger$} & \cmark & 90.5 & 89.5 & 91.5 & 81.8 \\
    
	\myalign{l}{\;\;\;\footnotesize $\rotatebox[origin=c]{180}{$\Lsh$}$ wo/ ASPP$^\dagger$} & \cmark & 90.1 & 89.0 & 91.1 & \hspace{-0.2em}\tabnode{\underline{86.0}} \\
	
    % \bottomrule
\end{tabular}

\begin{tikzpicture}[overlay]
\node[draw=dark_orange,fill=light_orange,fill opacity=.2,draw opacity=1,rounded corners = 1ex,fit=(1)(2),inner sep = 0pt,text opacity=1] {};
\end{tikzpicture}

    \caption{Results on the DAVIS 2016 validation set. CC denotes constant cost during the inference. $\dagger$ indicates YouTube-VOS is added during the training stage. $*$ denotes BL30K is added during the training stage. For both CC and non-CC methods, the best results are highlighted in \textbf{bold}, while the second best results are \underline{underlined}. FPS was averaged over 3 runs.}
    \label{table:davis16}
\end{table}

\begin{table}[!htbp]
    \centering
    % \begin{landscape}

\begin{tabular}{lccccc}
    \toprule
    \textbf{Method} & CC & $\JandF$ & $\J$ & $\F$ & FPS \\
    \midrule
    STM$^\dagger$z & \xmark & 81.8 & 79.2 & 84.3 & 10.2 \\
% 	KMN$^\dagger$~\cite{seong2020kernelizedMemory} & \xmark & 82.8 & 80.0 & 85.6 & $<$8.4 \\
% 	JOINT$^\dagger$~\cite{mao2021joint} & \xmark & 83.5 & 80.8 & 86.2 & 4.0 \\
% 	LCM$^\dagger$~\cite{hu2021learning} & \xmark & 83.5 & 80.5 & 86.5 & $<$8.5 \\
	RMNet$^\dagger$~\cite{xie2021efficient} & \xmark & 83.5 & 81.0 & 86.0 & $<$11.9 \\
	MiVOS$^{\dagger*}$~\cite{cheng2021mivos} & \xmark & 84.5 & 81.7 & 87.4 & 11.2 \\
% 	HMMN$^\dagger$~\cite{seong2021hierarchical} & \xmark & 84.7 & 81.9 & 87.5 & $<$10.0 \\
	STCN$^{\dagger*}$~\cite{cheng2021rethinking} & \xmark & 85.3 & 82.0 & 88.6 & \underline{20.2} \\
	GSFM~\cite{liu2022global} & \xmark & \underline{86.2} & 83.1 & 89.3 & $\approx$ 8.9 \\
	BATMAN~\cite{yu2022batman} & \xmark & \underline{86.2} & \underline{83.2} & 89.4 & - \\
	XMem$^{\dagger}$~\cite{cheng2022xmem} & \xmark & \underline{86.2} & 82.9 & \underline{89.5} & \textbf{22.6} \\
	XMem$^{\dagger*}$~\cite{cheng2022xmem} & \xmark & \textbf{87.7} & \textbf{84.0} & \textbf{91.4} & \textbf{22.6} \\
	\midrule
	GCNet~\cite{li2020fastGlobalContext} & \cmark & 71.4 & 69.3 & 73.5 & $<$25.0 \\
% 	Liang et al.~\cite{liang2020video} & \cmark & 74.6 & 73.0 & 76.1 & 4.0 \\
	%G-FRTM$^\dagger$~\cite{park2021learning} & \cmark & 76.4 & - & - & 18.2 \\
	PReMVOS~\cite{luiten2018premvos} & \cmark & 77.8 & 73.9 & 81.7 & 0.01 \\
	SwiftNet$^\dagger$~\cite{wang2021swiftnet} & \cmark & 81.1 & 78.3 & 83.9 & $<$25.0 \\
	SST~\cite{duke2021sstvos} & \cmark & 82.5 & 79.9 & 85.1 & - \\
% 	Ge et al.$^\dagger$~\cite{ge2021video} & \cmark & 82.7 & 80.2 & 85.3 & 6.7 \\
	RDE-VOS$^\dagger$~\cite{li2022recurrent} & \cmark & 84.2 & 80.8 & 87.5 & 27.0 \\
	RDE-VOS$^{\dagger*}$~\cite{li2022recurrent} & \cmark & \textbf{86.1} & \underline{82.1} & \textbf{90.0} & 27.0 \\
	
    & & & & & \\[-2ex]
	\myalign{l}{\hspace{-0.6em}\tabnode{MobileVOS}} & & & & & \\
	\myalign{l}{\;\;\;\footnotesize ResNet18$^\dagger$} & \cmark & 83.7 & 80.2 & 87.1 & \textbf{90.6} \\

	\myalign{l}{\;\;\;\footnotesize ResNet18$^{\dagger*}$} & \cmark & 85.0 & 81.7 & 88.3 & \textbf{90.6} \bfseries \\
	
	\myalign{l}{\;\;\;\footnotesize $\rotatebox[origin=c]{180}{$\Lsh$}$ model soup$^\dagger$} & \cmark & \underline{85.6} & \textbf{82.3} & \underline{88.9} & \textbf{90.6} \\
	
	\myalign{l}{\;\;\;\footnotesize MobileNetV2$^\dagger$} & \cmark & 82.2 & 78.7 & 85.7 & 79.1 \\

	\myalign{l}{\;\;\;\footnotesize $\rotatebox[origin=c]{180}{$\Lsh$}$ wo/ ASPP$^\dagger$} & \cmark & 81.8 & 78.3 & 85.3 & \hspace{-0.2em}\tabnode{\underline{81.3}} \\

\end{tabular}

\begin{tikzpicture}[overlay]
\node[draw=dark_orange,fill=light_orange,fill opacity=.2,draw opacity=1,rounded corners = 1ex,fit=(3)(4),inner sep = 0pt,text opacity=1] {};
\end{tikzpicture}

    \caption{Results on the DAVIS 2017 validation set. CC denotes constant cost during the inference.FPS was averaged over 3 runs.}
    \label{table:davis17}
\end{table}

\vspace{-1.2em}
\paragraph{YouTube-VOS}
Table~\ref{table:youtube} shows a comparison of our method against other state-of-the-art methods on the large-scale YouTube-VOS 2019 validation set. Our method outperforms RDE-VOS in the case where no BL30K pre-training is used and is competitive in the case where it is used. Unlike RDE-VOS, which trains with different loss weights for the YouTube evaluation to avoid overfitting on unseen classes, we report predictions using the same set of weights as in the DAVIS evaluation. We also observe only a $1.9$ $\JandF$ drop with respect to the STCN teacher model.

\begin{table}[!htbp]
    \centering
    \resizebox{1.\linewidth}{!}{\begin{tabular}{lcccccc}
    \toprule
    \textbf{Method} & CC & Overall & $\J_{seen}$ & $\F_{seen}$ & $\J_{unseen}$ & $\F_{unseen}$ \\
    \midrule
	STM$^{\dagger}$~\cite{oh2019video} & \xmark & 79.2 & 79.6 & 83.6 & 73.0 & 80.6 \\
	MiVOS$^{\dagger*}$~\cite{cheng2021mivos} & \xmark & 82.4 & 80.6 & 84.7 & 78.2 & \underline{85.9} \\
	STCN$^{\dagger*}$~\cite{cheng2021rethinking} & \xmark & \underline{84.2} & 82.6 & \textbf{87.0} & 79.4 & \textbf{87.7} \\
% 	GSFM$^{\dagger}$~\cite{liu2022global} & \cmark & - & - & - & - & - \\
	BATMAN$^{\dagger}$~\cite{yu2022batman} & \cmark & 84.5 & \textbf{89.3} & 79.0 & \underline{87.2} & 85.0 \\
	XMem$^{\dagger}*$~\cite{cheng2022xmem} & \cmark & \textbf{84.8} & \underline{89.2} & 80.3 & \textbf{88.8} & 85.8 \\
	\midrule
	CFBI$^{\dagger}$~\cite{yang2020collaborativeCFBI} & \cmark & 81.0 & 80.6 & 85.1 & 75.2 & 83.0 \\
	SST$^{\dagger}$~\cite{yang2020collaborativeCFBI} & \cmark & 81.8 & 80.9 & - & 76.6 & - \\
	SwiftNet$^{\dagger}$~\cite{wang2021swiftnet} & \cmark & 77.8 & 77.8 & 81.8 & 72.3 & 79.5 \\
	RDE-VOS$^{\dagger}$~\cite{li2022recurrent} & \cmark & 81.9 & 81.1 & 85.5 & 76.2 & 84.8 \\
	RDE-VOS$^{\dagger*}$~\cite{li2022recurrent} & \cmark & \textbf{83.3} & 81.9 & 86.3 & \textbf{78.0} & \textbf{86.9} \\

    & & & & & & \\[-2ex]
	\myalign{l}{\hspace{-0.6em}\tabnode{MobileVOS}} & & & & & & \\
	\myalign{l}{\;\;\;\footnotesize ResNet18$^{\dagger}$} & \cmark & 82.3 & 81.6 & 86.0 & 76.3 & 85.2 \\

	\myalign{l}{\;\;\;\footnotesize ResNet18$^{\dagger*}$} & \cmark & \underline{82.8} & \underline{82.1} & \underline{86.4} & \underline{77.0} & \underline{85.6} \\
	
	\myalign{l}{\;\;\;\footnotesize $\rotatebox[origin=c]{180}{$\Lsh$}$ model soup$^\dagger$} & \cmark & \textbf{83.3} & \textbf{83.2} & \textbf{87.7} & 76.9 & 85.3 \\
	
	\myalign{l}{\;\;\;\footnotesize MobileNetV2$^{\dagger}$} & \cmark & 80.3 & 80.4 & 84.6 & 74.0 & 82.4 \\

    \myalign{l}{\;\;\;\footnotesize $\rotatebox[origin=c]{180}{$\Lsh$}$ wo/ ASPP$^{\dagger}$} & \cmark & 80.1 & 79.0 & 83.2 & 75.1 & \tabnode{83.3} \\
    % \bottomrule
\end{tabular}

\begin{tikzpicture}[overlay]
\node[draw=dark_orange,fill=light_orange,fill opacity=.2,text opacity=1,rounded corners = 1ex,fit=(5)(6),inner sep = 0pt,text opacity=1] {};
\end{tikzpicture}
}

    \caption{Results on the YouTube-VOS 2019 validation set.}
    \label{table:youtube}
\end{table}

\paragraph{Qualitative results}
Figure \ref{fig:video_images} shows the segmentation of two identical models trained with and without knowledge distillation. In this example, we observe that the distilled model is able to successfully segment the panda despite undergoing drastically different views and occlusions.

\begin{table*}[htb]
\addtolength{\tabcolsep}{-0.4em}
% \addtolength{\tabrowsep}{-0.4em}
\centering
\begin{tabular}{ccccc}
\rotatebox{90}{\footnotesize \;\;\;\;\;\;\;\;$\mathcal{L}_{XE}\;\;only$} & \includegraphics[width=.23\linewidth]{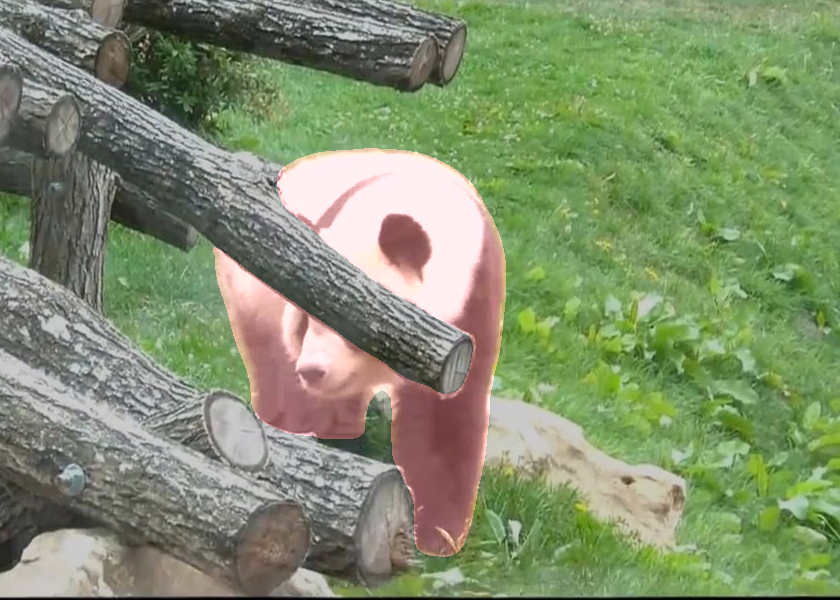} & \includegraphics[width=.23\linewidth]{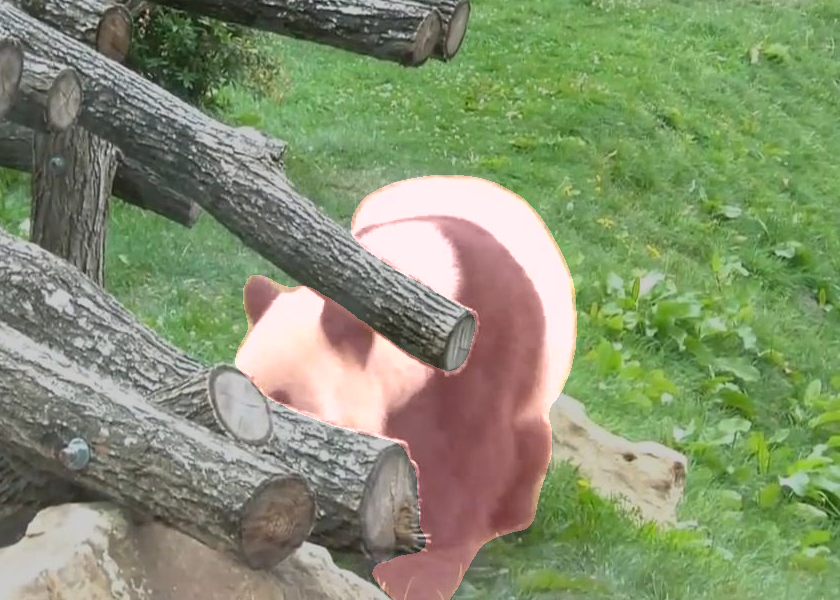} & \includegraphics[width=.23\linewidth]{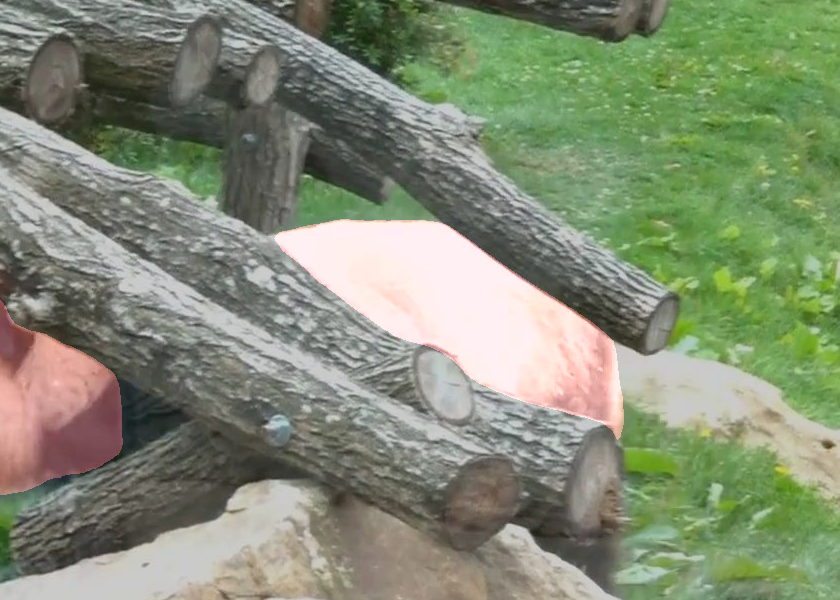} & \includegraphics[width=.23\linewidth]{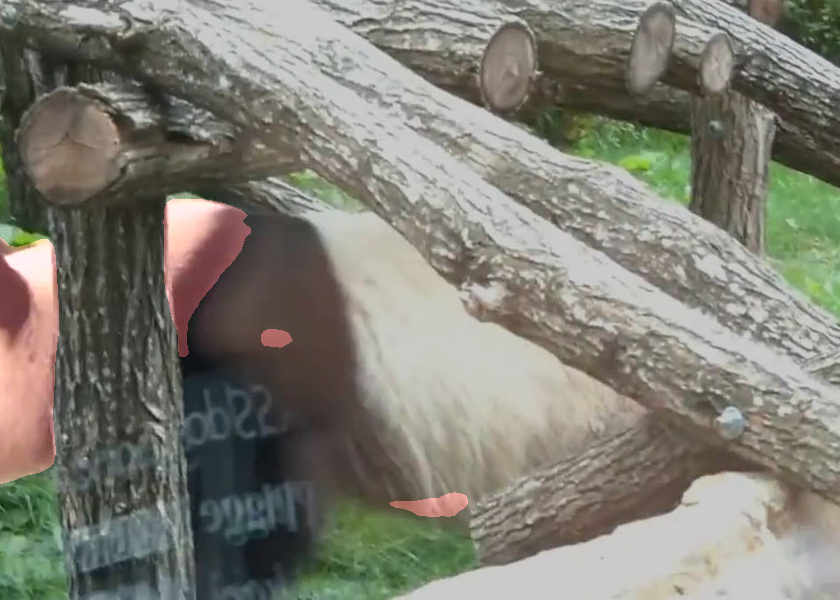} \\
\rotatebox{90}{\footnotesize $\mathcal{L}_{XE}+\mathcal{L}_{logit}+\mathcal{L}_{repr}$} & \includegraphics[width=.23\linewidth]{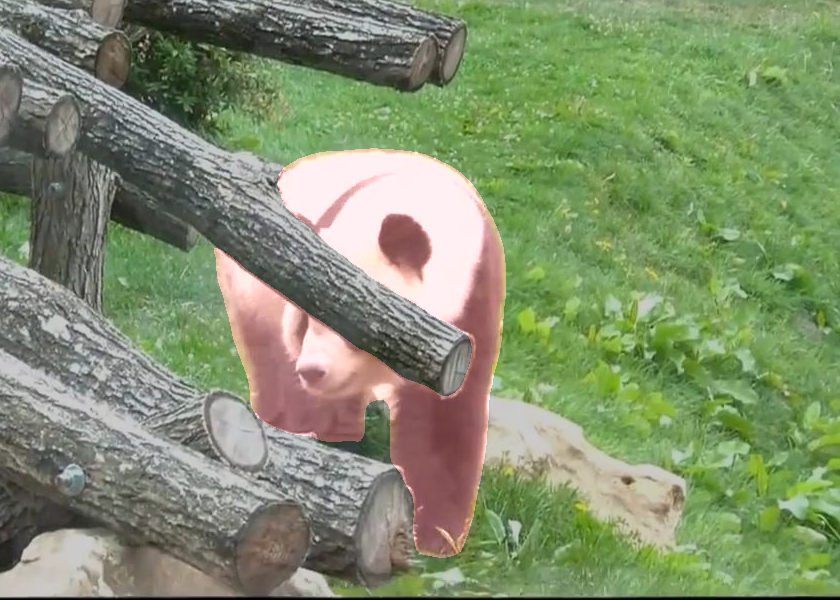} & \includegraphics[width=.23\linewidth]{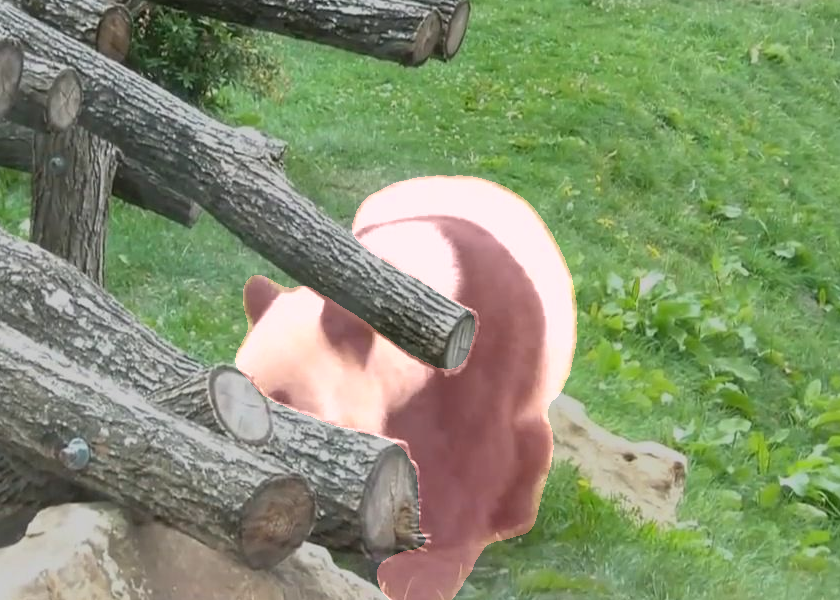} & \includegraphics[width=.23\linewidth]{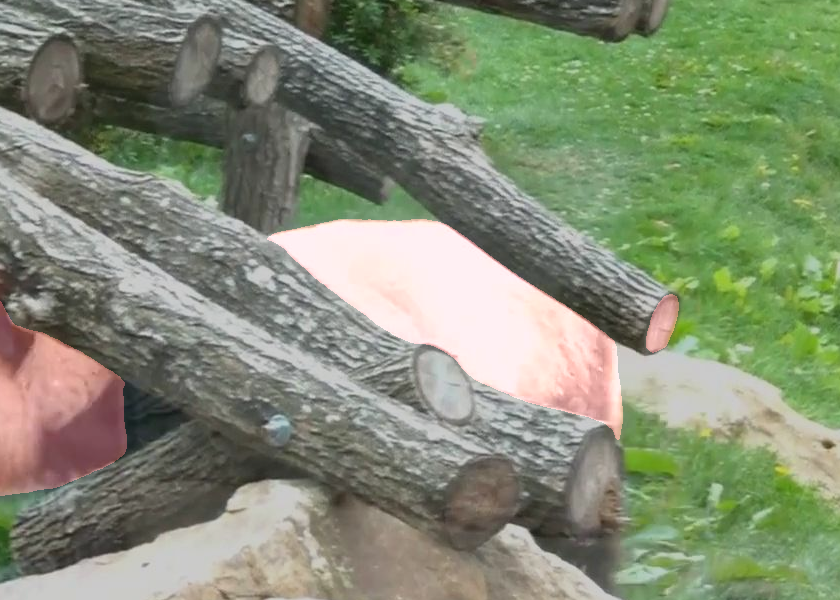} & \includegraphics[width=.23\linewidth]{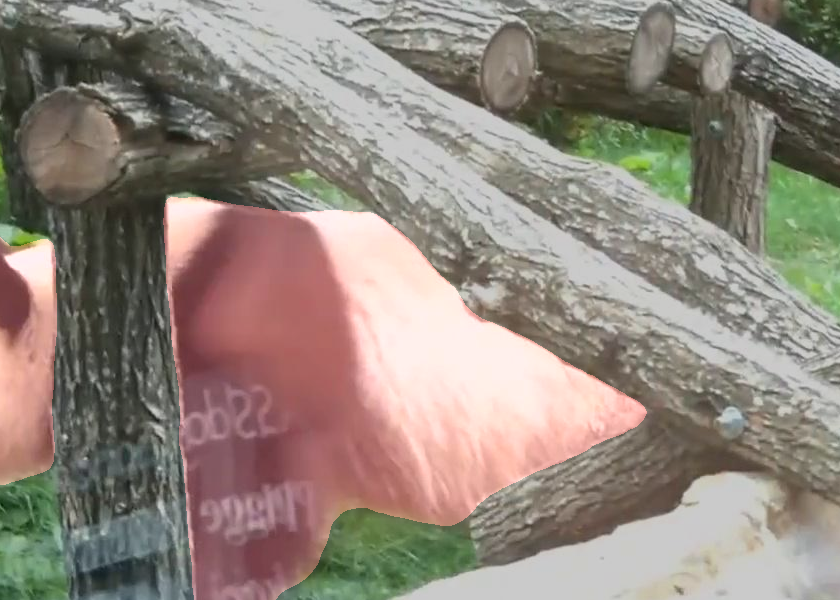} \\
\end{tabular}
\vspace{-0.5em}
\captionof{figure}{Comparing the segmentation of two models with and without constrastive learning or distillation being applied. Despite both models having finite memory, the cross-entropy only model fails to learn features which are consistent across all frames of the video. This is demonstrated in the last frame above, where the side of the panda is poorly segmented.}
\label{fig:video_images}
\end{table*}

\vspace{-1em}
\paragraph{Model size and inference latency}
To conduct a fair comparison of the computational cost of previous works, we jointly evaluate MobileVOS and other methods on the same set of hardware. We consider both a server-grade NVIDIA A40 and a desktop-grade NVIDIA 1080Ti. In all cases, we use the authors' provided code, but with a modification to accept randomly generated video sequences of two different durations. The results can be seen in Table \ref{table:performance_v2} and show that our MobileVOS models are significantly smaller, while retaining a constant low latency across both the short and long-duration videos. This result is attributed to the smaller backbones and the constant memory costs due to a finite memory queue length. Our largest model with ResNet18 has more than $3 \times$ the FPS than RDE-VOS but with $8 \times$ fewer parameters. Finally, we observe that the ResNet models perform better on the server-grade GPUs. This is in line with other results in the literature~\cite{Bianco_2018} and is related to the trade-off between model depth and GPU compute units available. In the following section, we show that we obtain a different outcome for mobile GPUs.

\begin{table}[!htbp]
    \centering
    \resizebox{1.\linewidth}{!}{
\centering
\begin{tabular}{lccccc}
    \toprule
    Method & Params(M) &     
    \multicolumn{2}{c}{\begin{tabular}{c}
        FPS \\
        \small\textit{NVIDIA A40}
    \end{tabular}} &
    \multicolumn{2}{c}{\begin{tabular}{c}
        FPS \\
        \small\textit{NVIDIA 1080Ti}
    \end{tabular}} \\
    \cmidrule{3-6}
    & & short & long (10$\times$) & short & long (10$\times$) \\
    \midrule
    
    STM~\cite{oh2019video} & 38.9 & 8.9 & 4.3 & 6.8 & \xmark \\
    
    GSFM~\cite{xie2021efficient} & 67.0 & 18.4 & 4.2 & 7.6 & \xmark \\
    
    STCN~\cite{cheng2021rethinking} & 54.4 & 37.4 & 8.3 & 18.1 & \xmark \\

    RDE-VOS~\cite{xie2021efficient} & 64.0 & 32.0 & 34.2 & 14.4 & 14.1 \\

    XMem~\cite{cheng2022xmem} & 62.2 & 38.6 & 39.9 & 12.6 & 12.7 \\
        
    \midrule
    
    MobileVOS &  &  \\
	\footnotesize ResNet18 & 8.1 & \textbf{144.7} & \textbf{145.4} & \textbf{76.0} & \textbf{76.3} \\
	
    \footnotesize MobileNetV2 & 2.5 & 99.9 & 99.1 & 61.6 & 60.6 \\

	\footnotesize $\rotatebox[origin=c]{180}{$\Lsh$}$ wo/ ASPP  & \textbf{1.9} & 105.1 & 103.4 & 66.8 & 67.4 \\
    \bottomrule
\end{tabular}}
    \caption{Performance evaluation on the same sets of hardware. The FPS metrics were evaluated on two randomly generated short and long video sequences with shape $480 \times 910$. The short videos consist of 50 frames, while the long videos consist of 500. In all cases, we use the authors' provided code and {\text{\ding{55}}} indicates that the models are exceeding the GPU memory limit.}
    \label{table:performance_v2}
\end{table}

\subsection{Mobile phone deployment}

\begin{figure}
  \centering %
  \includegraphics[width=.85\linewidth]{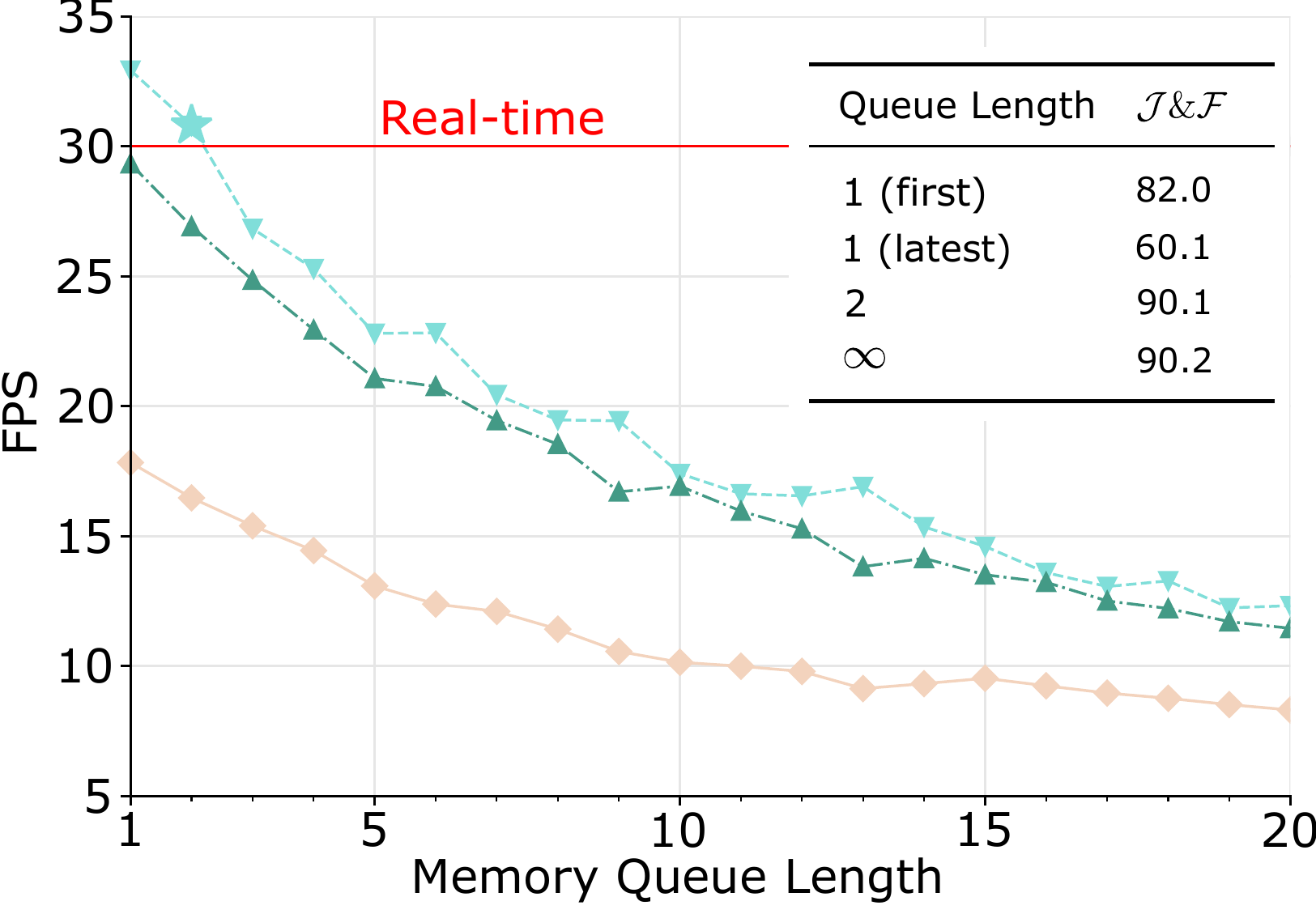}
  \caption{Runtimes of our proposed models with different memory queue lengths were evaluated on a Samsung Galaxy S22 GPU. The line with { \color[HTML]{80DED9} $\blacktriangledown$ markers} is the MobileNetV2 wo/ ASPP, { \color[HTML]{439A86} $\blacktriangle$ markers} is the MobileNetV2, while { \color[HTML]{F3D3BD} $\blacklozenge$ markers} are for the Resnet18. The table shows $\JandF$ on the DAVIS 2016 validation set for MobileNetV2 wo/ ASPP on memory queue lengths of 1, 2 and unbounded. We highlight the candidate that achieves the highest $\JandF$, while maintaining real-time performance with the star sign ({\color[HTML]{80DED9} $\star$}). The latency results are averaged across 100 runs.}
  \label{fig:mobile_perf}
\end{figure}

For the purpose of mobile deployment, we evaluate the inference latency of our models on the GPU of a Samsung Galaxy S22 device. The models are first converted to tflite format and then benchmarked using the official tflite benchmarking tool~\cite{tensorflow2015-whitepaper}. To validate our choice of fixed memory size with 2 entries, we show the model latency against queue sizes between 1 and 20 (see figure \ref{fig:mobile_perf}). MobileNetV2 wo/ ASPP model is the only model that is able to achieve real-time performance up to a memory queue length of 2. Figure \ref{fig:mobile_perf} also shows a comparison of the $\JandF$ performance on the DAVIS 2016 validation set at different memory queue lengths, where there is a minimal accuracy degradation by reducing this length to 2. These two points motivate the choice of a memory queue length of 2.

\vspace{-1em}
\paragraph{Model soups}
We consider the use of model soups~\cite{wortsman2022model} as a scheme for utilising the expressiveness of an ensemble without any additional inference costs. This is in contrast to multi-scale inference, which is typically adopted in the literature~\cite{cheng2021rethinking} and requires 4 forward passes. To do this, we firstly construct a set of models using the checkpoints from previous iterations trained with different values of $\omega$. Afterwards, we run a greedy selection using the training data as the validation metric. Tables \ref{table:davis17}, \ref{table:davis16}, and \ref{table:youtube} show these results, where we find that the model soup can surpass the performance of all the individual ingredients by at least 0.6 $\JandF$ on DAVIS 2017 and 0.5 $\JandF$ on YouTube, thus confirming the benefit of adopting this scheme in practice.

% \subsection{Conclusion and future work}
% \input{sections/conclusion}

%-------------------------------------------------------------------------

\vspace{-0.7em}
\section{Ablation Study}
\label{sec:ablation}
We now perform a thorough ablation study to highlight the benefit of our proposed distillation loss, and its extensions with boundary sampling and contrastive learning.

\vspace{-1em}
\paragraph{Loss terms.} We train our models with the two distillation losses $\mathcal{L}_{logit}$ and $\mathcal{L}_{repr}$ being incrementally added to the primary loss term (i.e. cross-entropy using ground-truth masks). The results presented in Table \ref{table:loss_ablation} show that both $\mathcal{L}_{logit}$ and $\mathcal{L}_{repr}$ introduce improvements.

\begin{table}[!htbp]
    \centering
    \resizebox{.9\columnwidth}{!}{%
    %                  DAVIS16        DAVIS17
% XE
% XE+Logits        90.91          83.49
% XE+Logits+Repr

\begin{tabular}{ccccc}
    \toprule
    \textbf{Losses} & $\JandF$ & $\J$ & $\F$ \\
    \midrule

    $\mathcal{L}_{XE}$ & 80.75 & 77.68 & 83.82 \\

	$\mathcal{L}_{XE}+\mathcal{L}_{logit}$ & 81.38 & 78.13 & 84.64 \\
	
	$\mathcal{L}_{XE}+\mathcal{L}_{logit}+\mathcal{L}_{repr}$ & \textbf{82.28} & \textbf{79.96} & \textbf{85.61} \\
	
    \bottomrule
\end{tabular} \label{tab:loss_ablation}
    }
    \caption{Evaluating the performance improvement on the YouTubeVOS dataset after introducing the two distillation losses. $\mathcal{J}$ and $\mathcal{F}$ metrics are averaged over both the seen and unseen classes.}
    \label{table:loss_ablation}
\end{table}

\vspace{-1.7em}
\paragraph{Interpolating between SCL and KD.} We sweep over various $\omega$ values in equation \ref{eq:omega} to interpolate between the supervised contrastive and distillation training regimes\footnote{All models are also trained with logit distillation.}. Table \ref{table:omega} shows the results using a ResNet18-backbone. The results show that the best results are achieved on DAVIS 2016 and 2017 with $\omega=1$, where the loss is reduced to the distillation loss. We see the opposite trend on YouTube; the best results are achieved with $\omega=0$, which reduces the loss to a supervised contrastive setting. Our hypothesis is that in the $\omega=0$ case, it is difficult to distill the unseen class knowledge from the teacher, whereas in $\omega=1$ scenario, the contrastive loss is robust to unseen classes due to a lowered dependence on object semantics.

\begin{table}
    \centering
    \resizebox{.7\columnwidth}{!}{%
    \begin{tabular}{cccc}
    \toprule
    % $\omega$ & $\JandF$ & $\J$ & $\F$ \\
    $\omega$ & DAVIS16 & DAVIS17 & YoutubeVOS \\
    % $\omega$ & DAVIS16 & DAVIS17 \\
    \midrule

    $0.0$ & \textbf{91.0} & 83.1 & \textbf{82.3} \\

    	$0.2$ & 90.6 & \underline{83.7} & 81.3 \\

    	$0.8$ & 90.2 & 83.4 & \underline{81.5} \\

    	$1.0$ & \underline{90.9} & \textbf{84.2} & 81.1 \\
     
    \bottomrule
\end{tabular} 
    }
    \caption{Interpolating between supervised-contrastive loss, $\omega = 0.0$, and (representation) distillation loss, $\omega = 1.0$. The performance improvement is robust across a range of values for $\omega$, which suggests the models jointly benefit from both training regimes. Different $\omega$ values offer flexibility to achieve better results.}
    \label{table:omega}
\end{table}

\begin{figure}[H]
    \centering
    \includegraphics[width=0.9\linewidth]{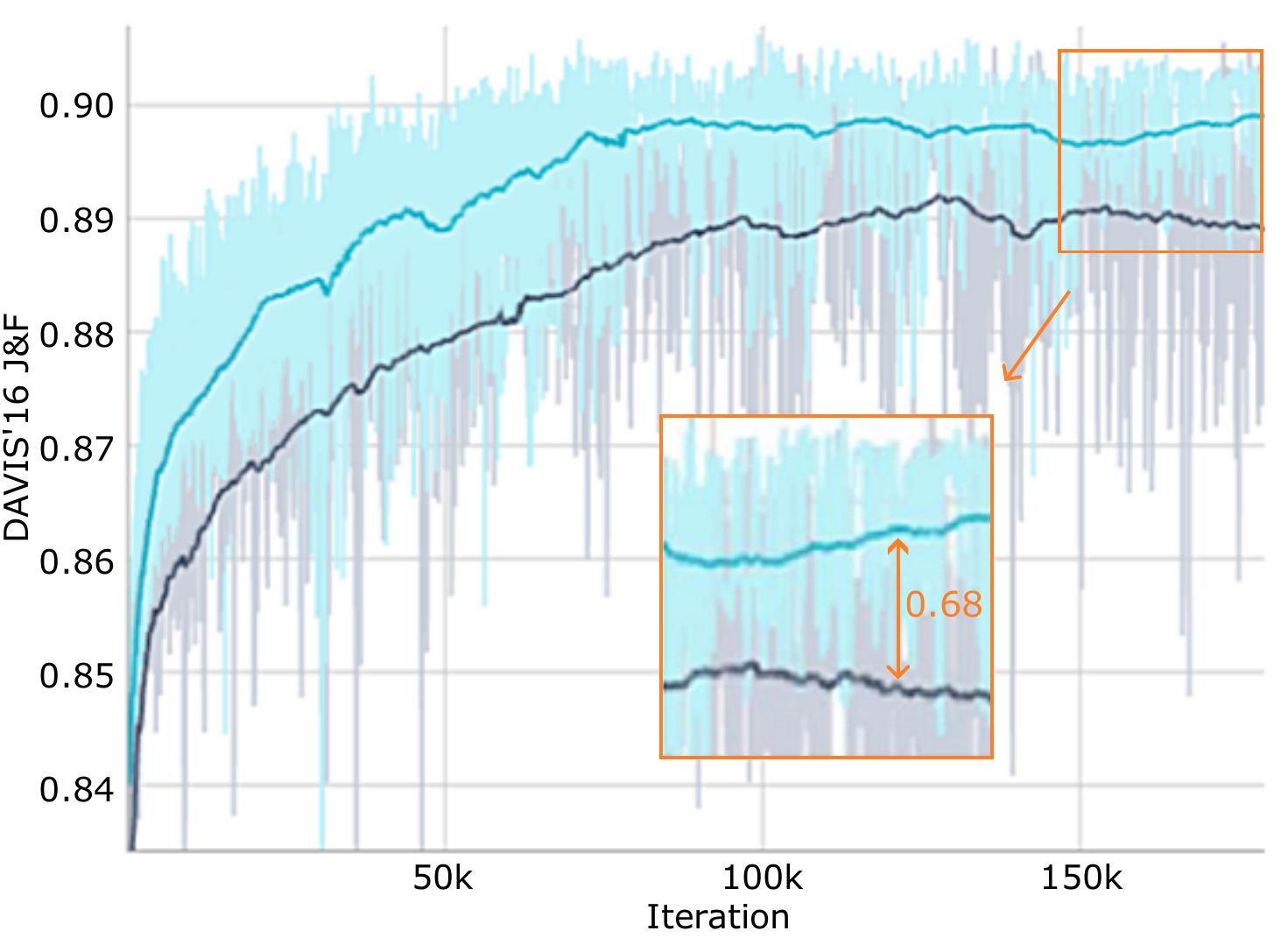}
    \caption{Validation accuracy for two models trained using random pixel and boundary pixel sampling strategies. Random sampling is shown in \textcolor{black}{black} and shows much slower convergence than boundary sampling, which is given in \textcolor{blue}{blue}.}
    \label{fig:pixel_sampling}
\end{figure}

\vspace{-1em}
\paragraph{Boundary sampling.} To demonstrate the performance improvement attributed to sampling boundary pixels, we perform an ablation experiment with a random sampling strategy. More specifically, we use logit distillation on all the pixels, and representation distillation on randomly selected pixels. Note that selecting all the pixels for representation distillation is prohibitively expensive due to the construction of the correlation matrices. The validation curves on DAVIS'16 for both strategies can be seen in figure~\ref{fig:pixel_sampling}. The curves show that sampling the boundary pixels yields an improved convergence rate. Furthermore, this faster convergence leads to better $\JandF$, as highlighted in the figure.

\section{Conclusion}
\label{sec:conclusion}
In this work, we present a simple loss that unifies knowledge distillation and contrastive learning. By applying this loss in the context of semi-supervised video object segmentation, we achieve competitive results with state-of-the-art at a fraction of the computational cost and model size. These models are compact enough to fit on a standard mobile device, while retaining real-time latency. 

%%%%%%%%% REFERENCES
{\small
\bibliographystyle{ieee_fullname}
\bibliography{references}
}

% added for arXiv
\newpage
\;
\newpage
% Removed for main paper submission
\section{Supplementary Material}
\subsection{Qualitative results}
We provide an extensive qualitative evaluation of both the ResNet and MobileNet variants. Firstly, we explore the performance of MobileVOS on an out-of-domain video and compare its predictions to XMem, whereby we observe that, although some segmentation errors occur, they are far less significant. Secondly, we provide real-time predictions of a single object, given on a mobile device. We expose this object to severe occlusions to highlight the robustness of our models in the wild. Finally, we show the generality of SVOS in its application with video inpainting.

\begin{figure*}[htb]
    \centering %
    
\begin{subfigure}{0.24\textwidth}
\centering
  First frame
\end{subfigure}\hfil %
\begin{subfigure}{0.24\textwidth}
\centering
  1. Similar features
\end{subfigure}\hfil %
\begin{subfigure}{0.24\textwidth}
\centering
  2. Shot changes
\end{subfigure}\hfil %
\begin{subfigure}{0.24\textwidth}
\centering
  3. Drift
\end{subfigure}
\medskip
\begin{subfigure}{0.24\textwidth}
  \includegraphics[width=\linewidth]{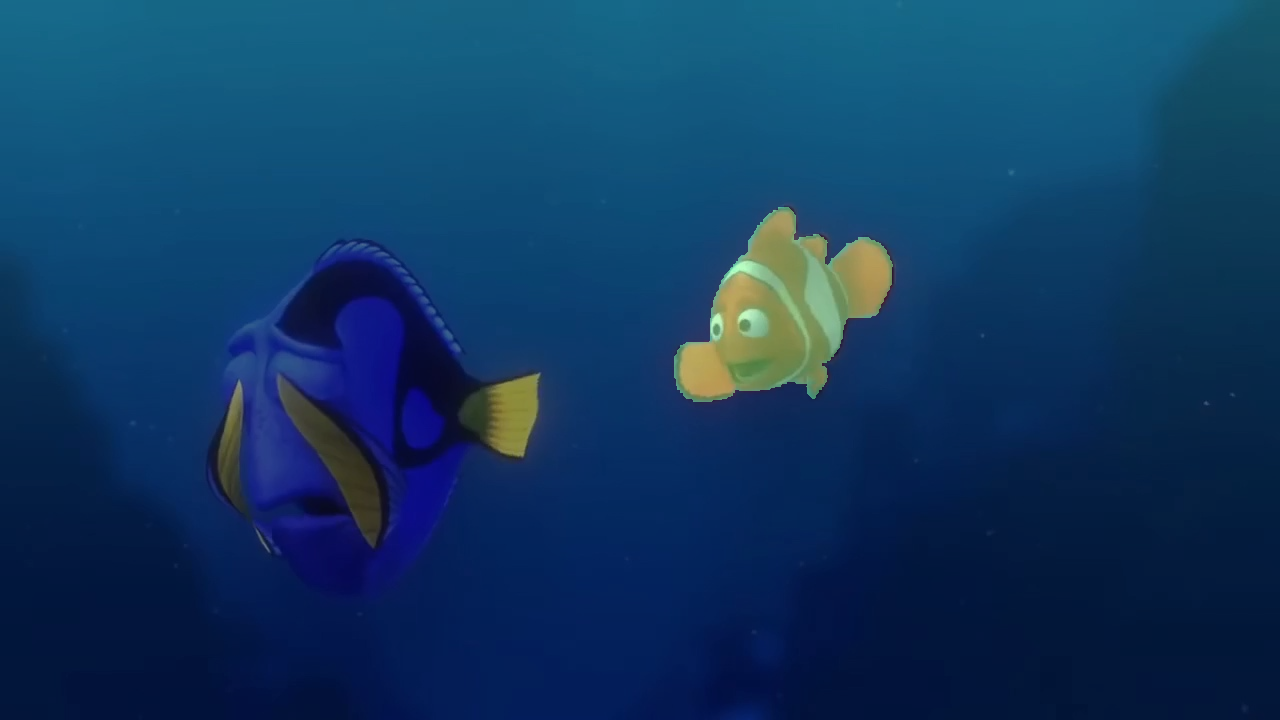}
\end{subfigure}\hfil %
\begin{subfigure}{0.24\textwidth}
  \includegraphics[width=\linewidth]{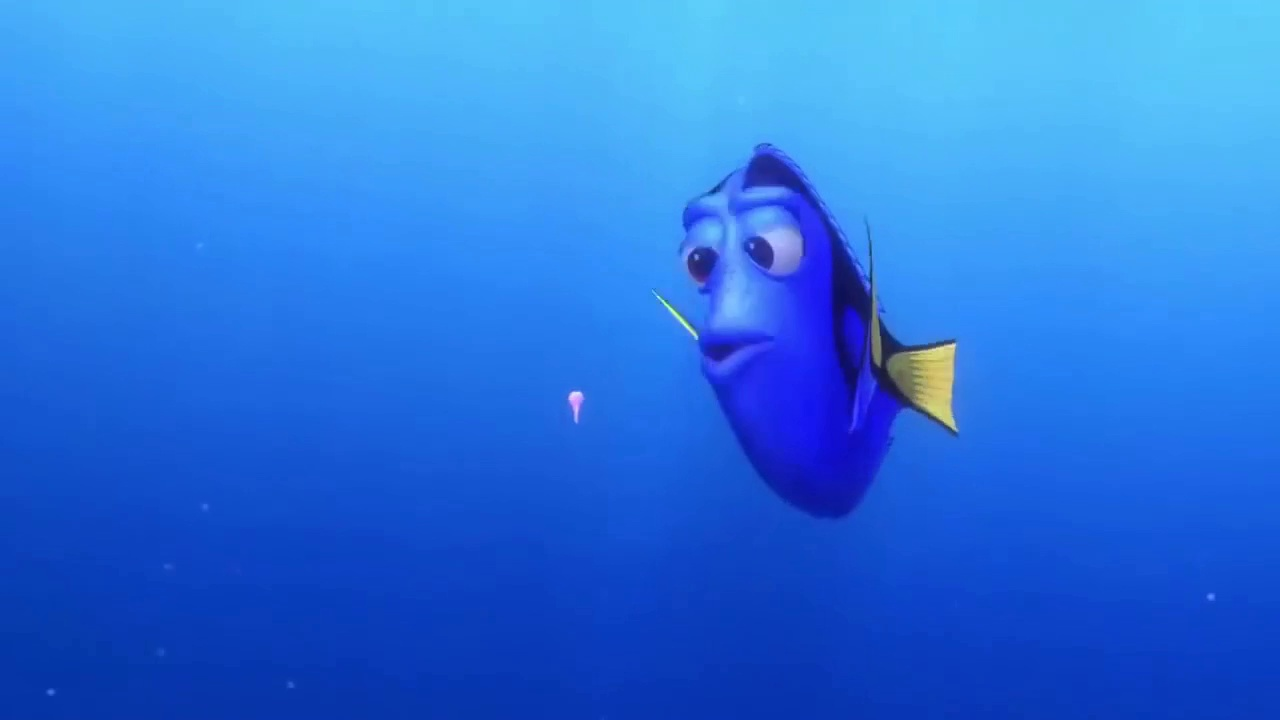}
\end{subfigure}\hfil %
\begin{subfigure}{0.24\textwidth}
  \includegraphics[width=\linewidth]{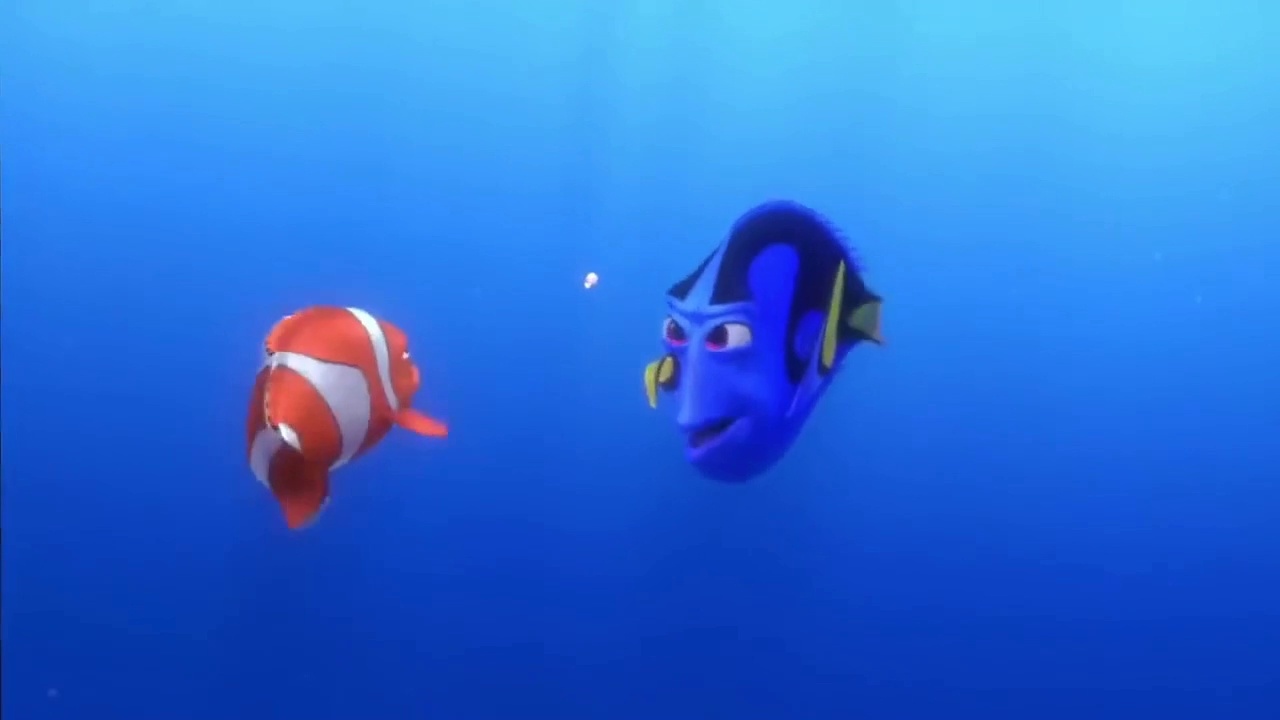}
\end{subfigure}\hfil %
\begin{subfigure}{0.24\textwidth}
  \includegraphics[width=\linewidth]{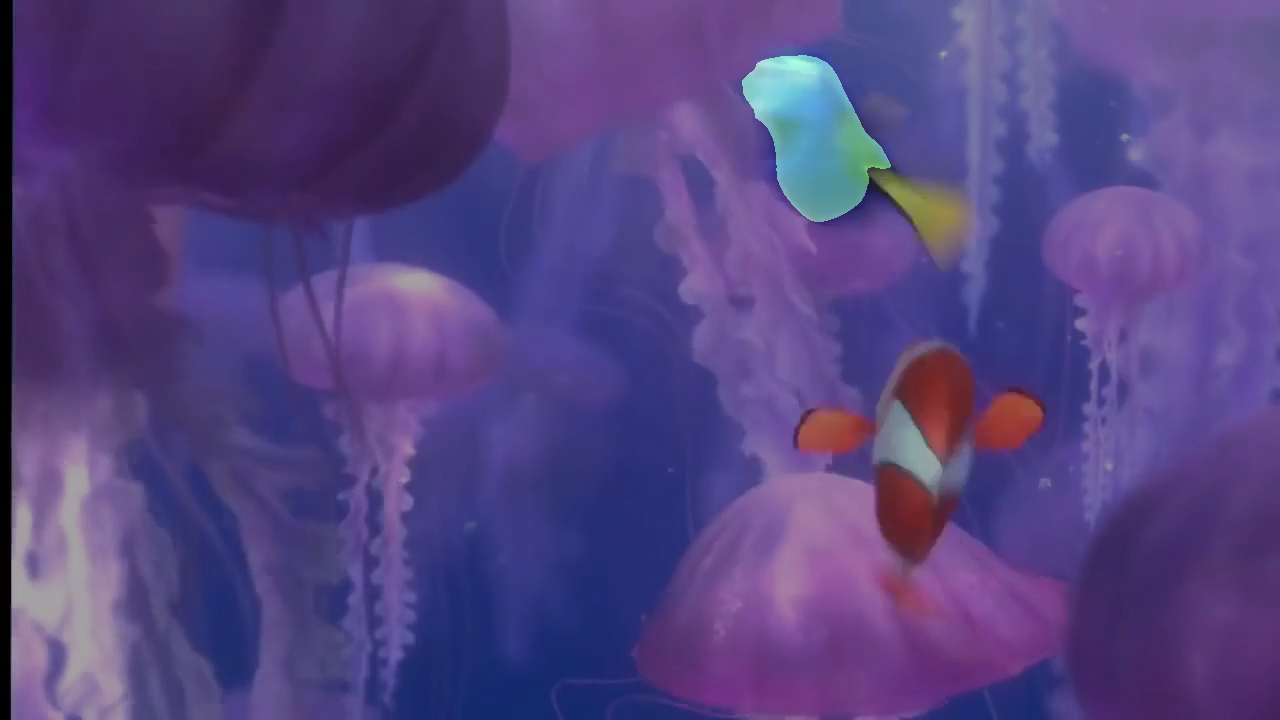}
\end{subfigure}
\medskip
\begin{subfigure}{0.24\textwidth}
  \includegraphics[width=\linewidth]{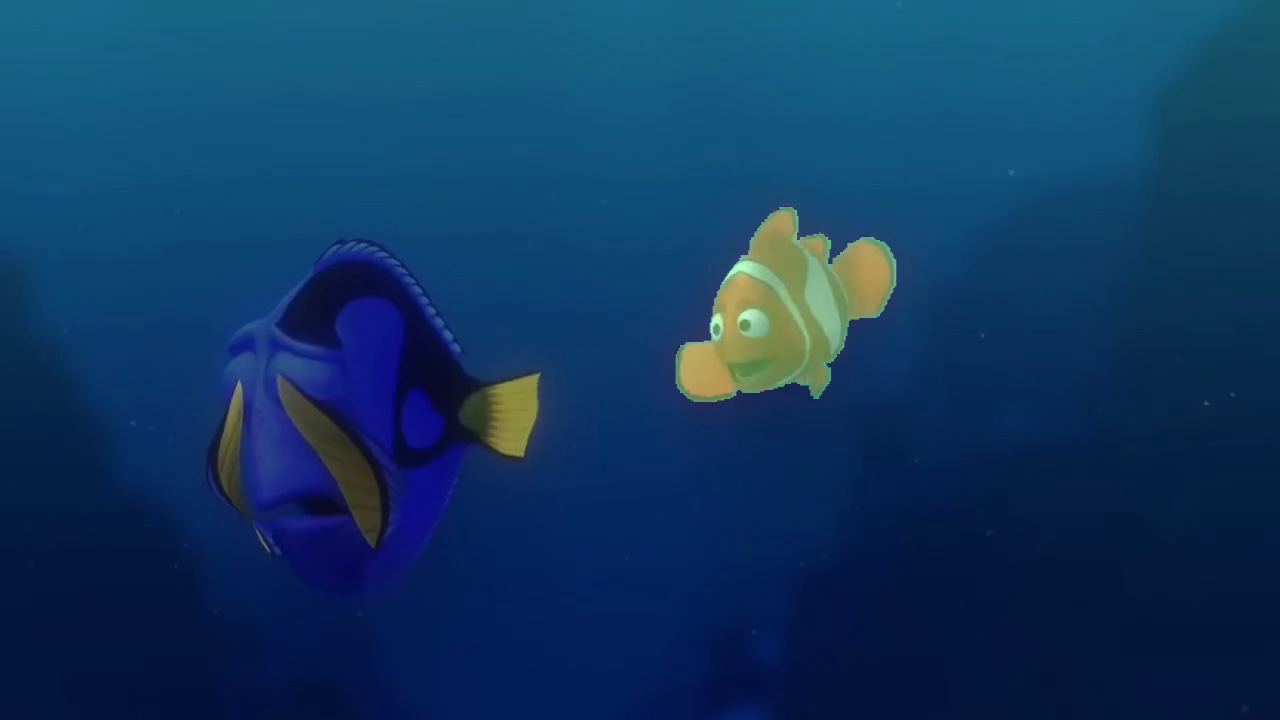}
\end{subfigure}\hfil %
\begin{subfigure}{0.24\textwidth}
  \includegraphics[width=\linewidth]{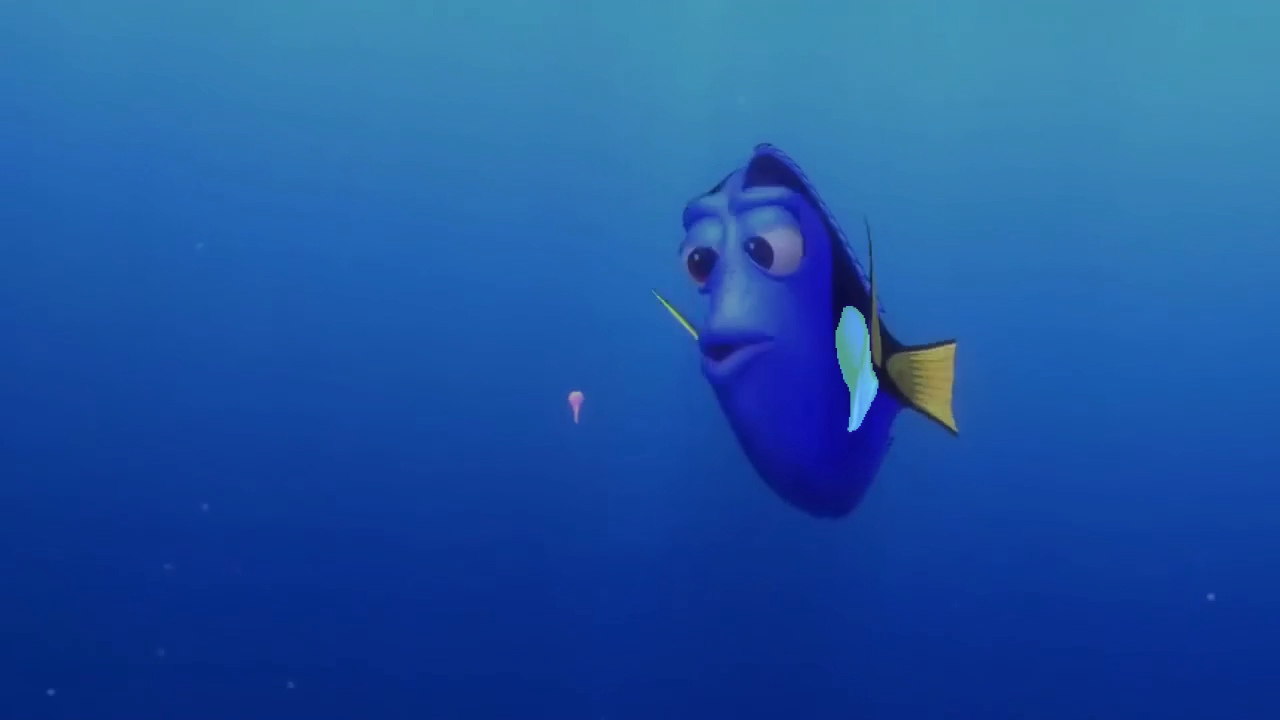}
\end{subfigure}\hfil %
\begin{subfigure}{0.24\textwidth}
  \includegraphics[width=\linewidth]{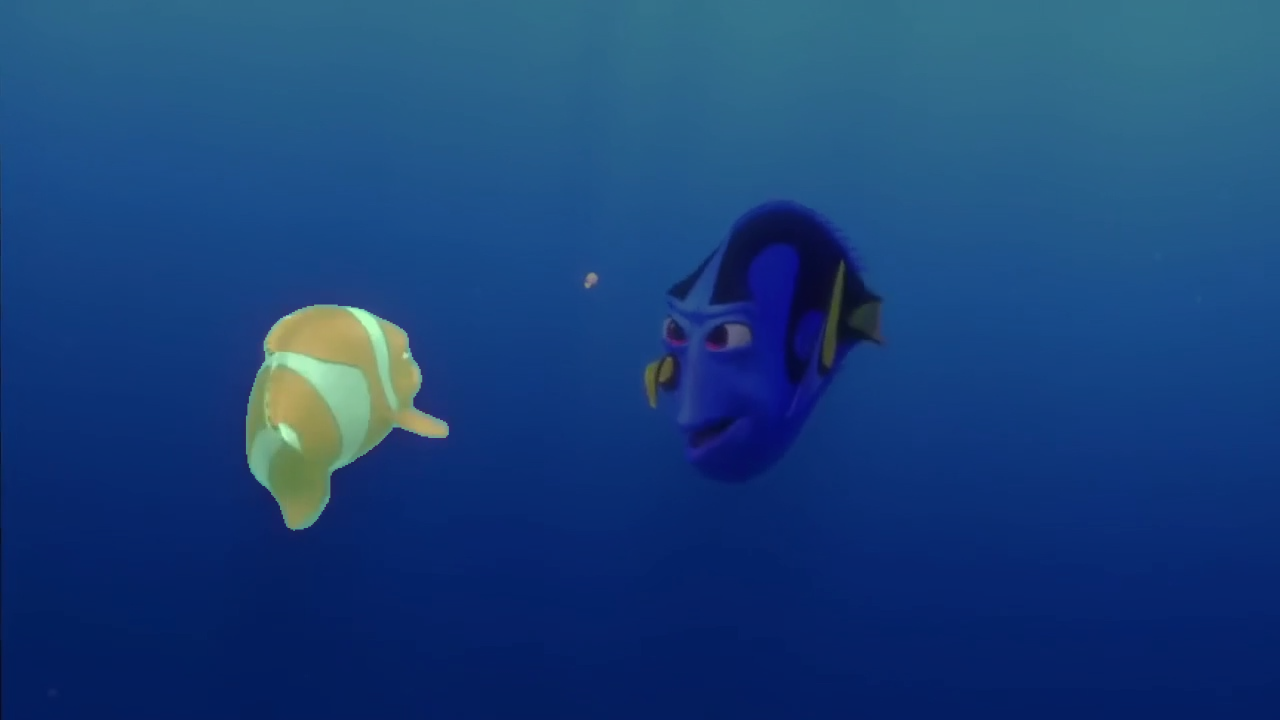}
\end{subfigure}\hfil %
\begin{subfigure}{0.24\textwidth}
  \includegraphics[width=\linewidth]{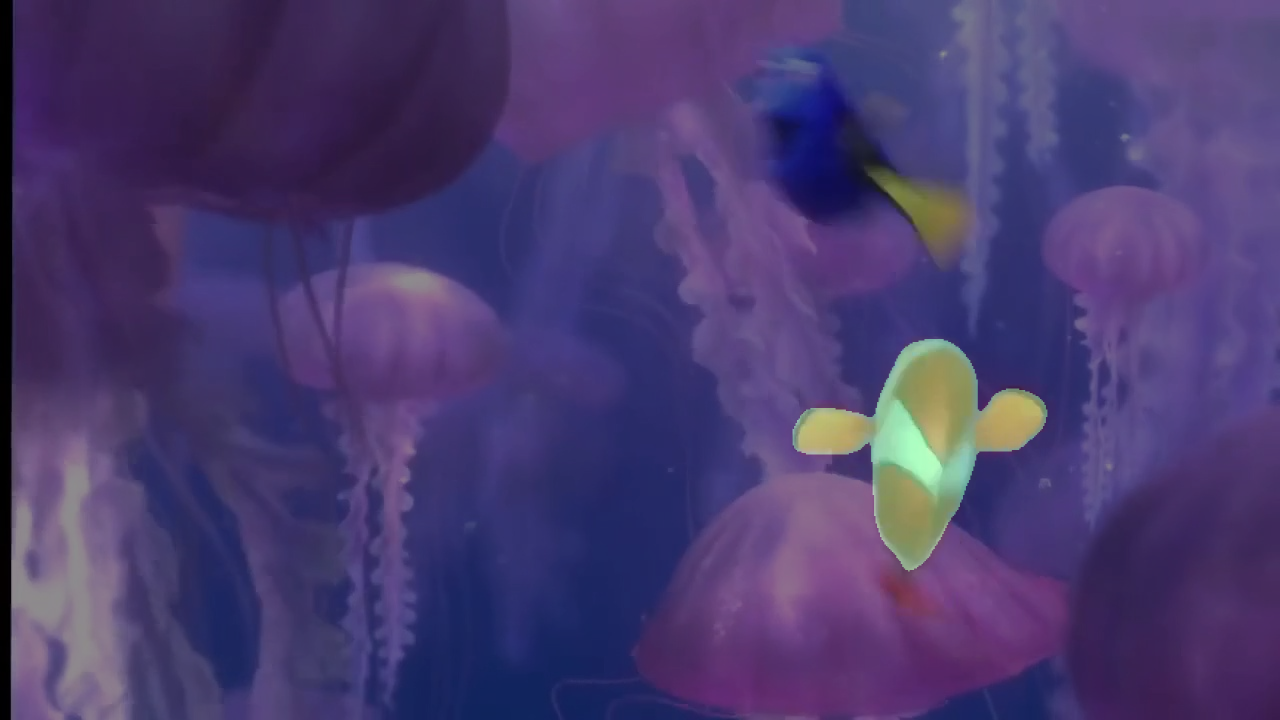}
\end{subfigure}
\vspace{-1em}
\caption{Comparison to XMem on a long out-of-domain single object segmentation task. Top row are the predictions by XMem, while the bottom row is from the ResNet MobileVOS. We categorise and highlight 3 distinct segmentation errors that can occur. The colour shift is just an artifact of how the masks are overlayed on the frames and is unrelated to the segmentation.}
\vspace{-.2em}
\label{fig:out_of_domain}
\end{figure*}

\vspace{-0.5em}
\paragraph{Out-of-domain}
The mask predictions given in figure \ref{fig:out_of_domain} demonstrate the robustness of MobileVOS to domain shifts, unseen classes, and camera shot changes. We compare these predictions to those given by XMem and observe 3 distinct failure modes that are unique to each of these model, where these failure modes are tied to the underlying architectures and memory models used. Although some segmentation errors occur only on MobileVOS, and not XMem, we expect that this is simply a trade-off imposed by the smaller network capacity, and the other types of failure modes (observed only in XMem) are much more detrimental.
\begin{enumerate}
    \item \textbf{Similar features} The second frame shows some poor segmentation on the wrong object, which we attribute to the smaller network capacity that is unable to learn sufficiently discriminative features. This is not observed in XMem due to the much larger backbones.
    \item \textbf{Shot changes} XMem can fail to segment the correct objects under camera shot changes since the model is matching features to a long sequence of intermediate frames which do not include the main object.
    \item \textbf{Drift} After XMem makes this first mistake, the model then begins to drift. MobileVOS does not suffer from this problem due to only storing the first and most recent frames/masks in memory. This drift leads to XMem poorly segmenting later frames in the video, including segmenting the wrong object.
\end{enumerate}

We have included the full length videos alongside this supplementary document. This video example highlights limitations of the YouTube and DAVIS evaluation datasets, which do not consider domain shifts or camera shot changes.

\begin{table}[h]
    \centering
    \begin{tabular}{cccc}
        \toprule
        Poly Loss & DAVIS 2016 & DAVIS 2017 \\
        \midrule
        \textbf{with}& 89.8 & 80.1 \\
        \textbf{without} & 89.8 & 79.9 \\
        \bottomrule
    \end{tabular} \label{tab:poly_loss}
    \caption{Evaluating the impact of the poly loss on both the DAVIS 2016 and DAVIS 2017 datasets, where $\epsilon = 1$ and the query encoder uses a MobileNetV2 backbone wo/ ASPP.}
    \label{table:performance_polyloss}
\end{table}

\begin{figure}[h]
    \centering
    \includegraphics[width=0.7\linewidth]{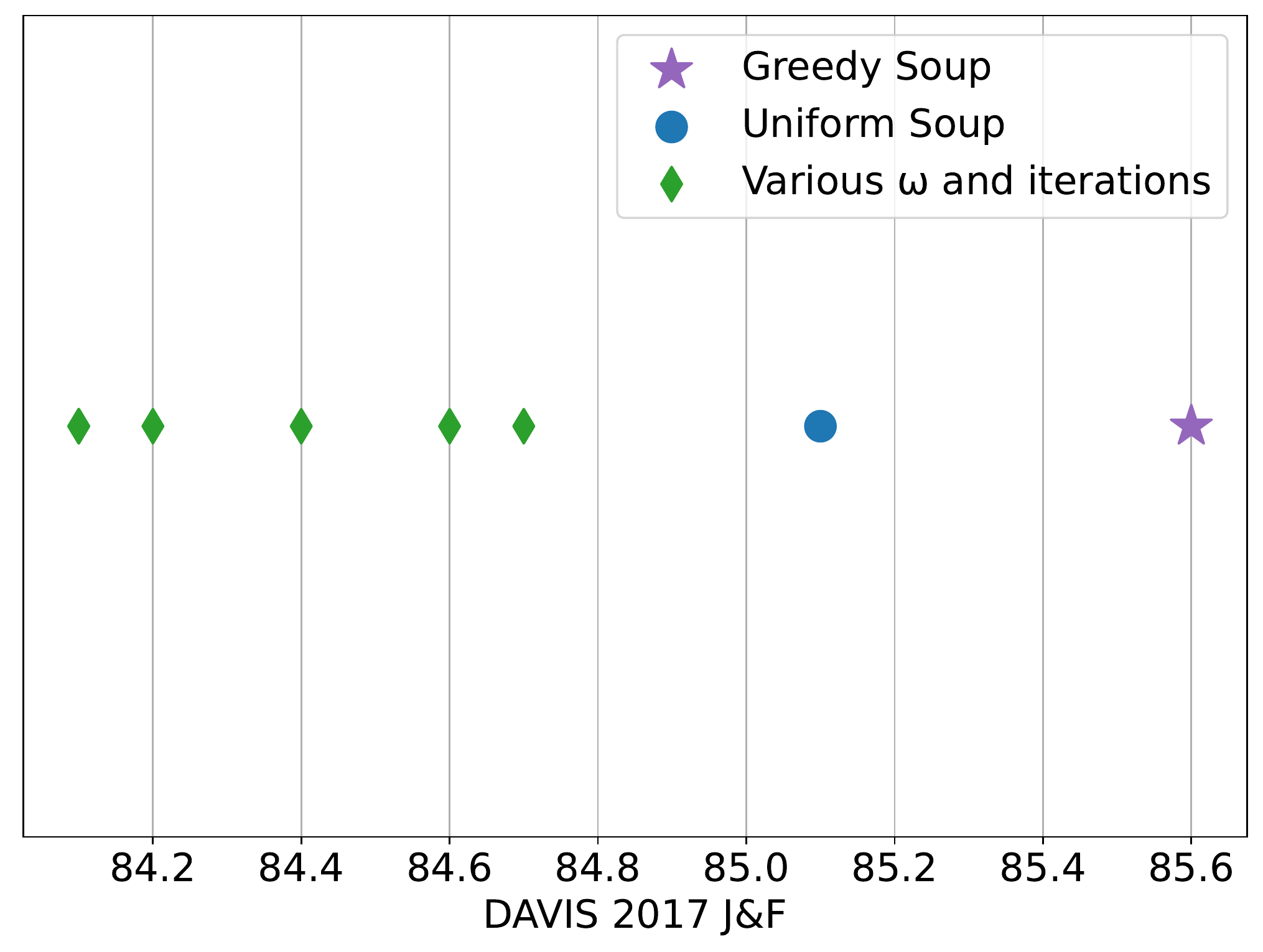}
    \caption{Model soups can improve the accuracy on both the DAVIS and YouTube datasets, without any additional inference costs.}
    \label{fig:model_soup}
\end{figure}

\vspace{-0.2em}
\paragraph{On-device Long term occlusion}
We demonstrate the robustness of our most efficient MobileVOS model (MobileNet V2 wo/ ASPP) under severe occlusion in real-time and on a mobile device. The results can be seen in the attached OnDeviceOcclusion.mp4 video and show almost no segmentation errors.

\begin{figure*}
    \centering
    \includegraphics[width=0.9\linewidth]{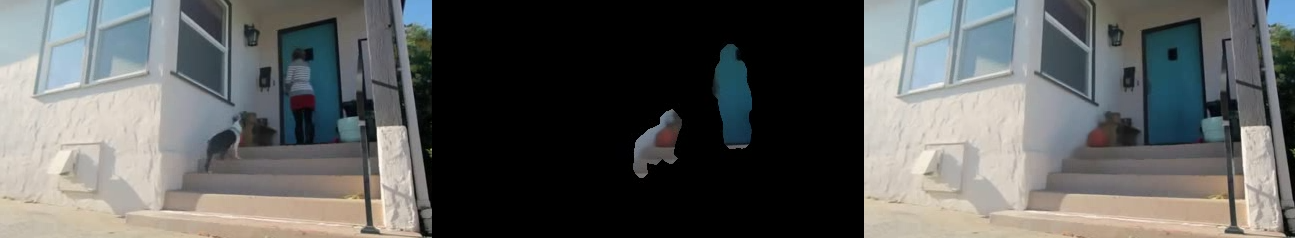}
    \caption{Using MobileVOS in conjunction with video in-painting to remove selecting objects.}
    \label{fig:inpainting}
\end{figure*}

\vspace{-0.5em}
\paragraph{Image inpainting}
A practical use case of SVOS is video inpainting. This task often requires per-frame masks, which indicate the areas to be inpainted. We use the MobileVOS ResNet18 variant to generate these per-frame masks, and then perform inpainting using FuseFormer~\cite{liu2021fuseformer}. We use the original operating resolution of FuseFormer (240p), where the segmentation masks are downsized to this resolution accordingly. In videos with multiple objects, we merge the masks into a single binary mask and use this as the inpainting input. We show an example of this application on a video provided in the YouTube validation split (see the attached video or figure \ref{fig:inpainting}). Despite significant object movement, our model is still able to provide highly accurate per-frame masks that lead to visually appealing inpainting results.

\subsection{Poly loss ablation}
Table \ref{table:performance_polyloss} shows the $\mathcal{J}\&\mathcal{F}$ results on both the DAVIS 2016 and DAVIS 2017 validation splits with and without using the additional poly loss component. In these experiments, we train the MobileNet (wo/ ASPP) backbone with no distillation or contrastive learning and observe that the poly loss can be safely removed without impacting the models performance.

\subsection{Model soups}
We use models soups~\cite{wortsman2022model} as an alternative to multi-scale inference, which is typically adopted in the SVOS literature~\cite{cheng2021rethinking}. Unlike model soups, multi-scale inference can incur significant additional inference costs due to multiple forward passes at different resolutions. Figure \ref{fig:model_soup} shows the accuracy of a few checkpoints with varying values of $\omega$ and at different iterations of training. By simply averaging the weights of all of these models, we achieve a significant increase in the DAVIS 2017 validation accuracy. However, by adopting a greedy selection process, we are able to achieve a much more significant increase. One noticeable observation from this process is that no additional data is needed for selecting the model to be included in the soup - they are simply conditionally added based on the observed training accuracy.

\vspace{-0.5em}
\subsection{Background - Kernel Perspective}
\Renyi's $\alpha$-entropy~\cite{Renyi1960OnInformation} of order $\alpha \in (0, 1) \cup (1, \infty)$ 
provides a natural extension of Shannon's entropy. Consider a random variable $X$ with probability density function (PDF) $f(x)$ in a finite set $\rchi$, the $\alpha$-entropy $\textbf{H}_{\alpha}(X)$ is defined as:
\begin{align}
    \textbf{H}_{\alpha}(f) = \frac{1}{1 - \alpha} \log_2 \int_{\chi} f^{\alpha}(x) dx
\end{align}
Where the limit as $\alpha\rightarrow 1$ is the well-known Shannon entropy. \cite{SanchezGiraldo2015MeasuresKernels, Williams2010NonnegativeInformation} propose a set of quantities that closely resemble \Renyi's entropy and omit the need for evaluating the underlying probability distributions. These information quantities are estimated directly from the data and are based on the theory of infinitely divisible matrices. Their usage leverages the representational power of reproducing kernel Hilbert spaces (RKHS), which is a concept that has been widely studied and adopted in classical machine learning. These estimators have been successfully applied in the context of knowledge distillation for image classification, reading comprehension, and binary network classification~\cite{Miles2021InformationDistillationBMVC}.

For completeness, we now provide definitions of these entropy-based quantities and their connections with positive semidefinite matrices. This idea then leads to a multi-variate extension using Hadamard products, from which conditional and mutual information can be defined. For brevity, we omit the proofs and connections with \Renyi's axioms, which can be found in \cite{SanchezGiraldo2015MeasuresKernels, Williams2010NonnegativeInformation}.\par
\textit{Definition 1}: Let $X = \{x^{(1)}, \dots x^{(n)}\}$ be a set of $n$ data points of dimension $d$ and $\kappa : X \times X \rightarrow \Real$ be a real-valued positive definite kernel. The Gram matrix $\mathbf{K}$ is obtained from evaluating $\kappa$ on all pairs of examples, that is $K_{ij} = \kappa(x^i, x^j)$. The matrix-based analogue to \Renyi’s $\alpha$-entropy for a normalized positive definite (NPD) matrix $\mathbf{A}$ such that $tr(\mathbf{A}) = 1$, can be given by the following functional: 
\begin{align}
    \mathbf{S}_{\alpha}(\mathbf{A}) &= \frac{1}{1 - \alpha}\log_{2}(tr(\mathbf{A}^{\alpha})) \\
    &= \frac{1}{1 - \alpha}\log_{2}\left[\sum_{i=1}^{n}\lambda_{i}(\mathbf{A}^{\alpha})\right]
    \label{eqn:entropy}
\end{align}
where $\mathbf{A}$ is the kernel matrix $\mathbf{K}$ normalised to have a trace of $1$ and $\lambda_i(\mathbf{A})$ denotes its $i$-th eigenvalue.
This estimator can be seen as a statistic on the space computed by the kernel $\kappa$, while also satisfying useful properties attributed to entropy. 

\textit{Definition 2}: Let $X$ and $Y$ be two sets of data points. After computing the corresponding Gram matrices $\mathbf{A}$ and $\mathbf{B}$, the joint entropy is then given by:
\begin{align}
    \textbf{S}_{\alpha}(\mathbf{A}, \mathbf{B}) = \textbf{S}_{\alpha}\left(\frac{\mathbf{A} \circ \mathbf{B}}{tr(\mathbf{A} \circ \mathbf{B})}\right)
    \label{eqn:joint_entropy}
\end{align}
where $\circ$ denotes the Hadamard product between two matrices. Using these two definitions, the notion of conditional entropy and mutual information can be derived. We focus on the mutual information, which is given by:
\begin{align}
    \textbf{I}_{\alpha}(\mathbf{A} ; \mathbf{B}) = \textbf{S}_{\alpha}(\mathbf{A}) + \textbf{S}_{\alpha}(\mathbf{B}) - \textbf{S}_{\alpha}(\mathbf{A}, \mathbf{B})
    \label{eqn:mutual_info}
\end{align}

\subsection{Decomposing the representation loss}
In this section, we provide an intricate connection between the proposed loss, mutual information, and contrastive learning. By bridging between these two training regimes, we find that models can benefit from minimising a linear weighting of these two objectives. We hope that this abstract lense can provide additional insights into the training dynamics for learning, and specifically in the context of very practical dense prediction tasks.

% \begin{theorem}
% \footnote{Up to a scale factor.}
\paragraph{Relating $\mathcal{L}_{repr}$ to mutual information}
\textit{More formally, in the case where $\omega = 1$, we show that minimising $\mathcal{L}_{repr}$ is equivalent to maximising the pixel-wise mutual information between the student and teacher representations.}
% \end{theorem}

% \begin{lemma}
Given $\mathbf{A}$ is a real symmetric matrix, then $\norm{\mathbf{A}}_F^2 = tr(\mathbf{A}\mathbf{A}^T) = \sum_{i=1}^{n} \lambda_i(\mathbf{A^2})$. This follows from the definition of the Frobenius norm of a matrix, $\norm{\mathbf{A}}_F^2 = \sum_{ij} \mathbf{A}_{ij}^2 $. \\ 
The trace term can be expanded as follows $tr(\mathbf{A}\mathbf{A}^T) = \sum_{i} \left(\mathbf{A}\mathbf{A}^T\right)_{ii} = \sum_{i} \sum_{j}\mathbf{A}_{ij}\mathbf{A}_{ji}$. Since $\mathbf{A}$ is symmetric, $\mathbf{A}_{ij} = \mathbf{A}_{ji}$ and thus $tr(\mathbf{A}\mathbf{A}^T) = \sum_{i} \sum_{j}\mathbf{A}_{ij}^2 = \norm{\mathbf{A}}_F^2$. Finally, the equality between the trace of a matrix and the sum of eigenvalues is a known relation in linear algebra.
% \end{lemma}

The representations, $\mathbf{Z}_S$ and $\mathbf{Z}_T$, are $L2$ normalised and thus the correlation matrices $\mathbf{C}$ will have $1$s along their leading diagonal. These matrices are real and symmetric, which allows use to use the relation derived above. 

The representation loss $\mathcal{L}_{repr}$ can be decomposed into the difference of two information-theoretic quantities, namely the entropy and joint entropy.

\begin{align}
    \mathcal{L}_{repr} &= \frac{1}{\mid \mathbf{C}_s \mid} \bigg( \log_2 {\norm{\mathbf{C}_s}^2} - \log_2 {\norm{\mathbf{C}_{s} \odot \mathbf{C}_{t}}^2} \bigg) \label{eq:1} \\
    &= \frac{1}{\mid \mathbf{C}_s \mid} \bigg( -\mathbf{S}_2(\mathbf{Z}_S) + \mathbf{S}_2(\mathbf{Z}_S ; \mathbf{Z}_T)\bigg) \label{eq:2}
\end{align}

Equation \ref{eq:2} follows from \ref{eq:1} using the definitions for the entropy estimators in equation \ref{eqn:entropy} and \ref{eqn:joint_entropy} with $\alpha = 2$. Maximising the mutual information can be given as follows:

\begin{align}
    \mathcal{L}_{mi} &= -\mathbf{I}_{2}(\mathbf{Z}_S ; \mathbf{Z}_T) \\
    &= \cancel{-\mathbf{S}_{2}(\mathbf{Z}_T)} - \mathbf{S}_{2}(\mathbf{Z}_S) + \mathbf{S}_{2}(\mathbf{Z}_S ; \mathbf{Z}_T)
\end{align}

Where the first entropy term can be omitted since no gradients flow through the teachers representation. From an optimisation perspective, these two losses are then equivalent. The only distinction is in the pixel-wise sampling strategy, where we opt to select only the boundary pixels, which leads to much faster model convergence.

\paragraph{Relating $\mathcal{L}_{repr}$ to contrastive learning} \textit{In the case where $\omega = 0$, minimising $\mathcal{L}_{repr}$ is a pixel-wise contrastive objective.}

The loss can be deconstructed into the sum of positive and negative pixel-wise pairings. In the case where $\omega = 0$, $\mathcal{C}_{ty} = \mathcal{C}_{y} = \mathbf{Y}\mathbf{Y}^T$.

\begin{align}
    \mathbf{C}_{y} = \left(\mathbf{Y}\mathbf{Y}^T\right)_{ij} = \begin{cases} 
      1 & j \in \mathcal{P}_i \\
      0 & j \in \mathcal{N}_i \\
   \end{cases}
\end{align}

where $\mathcal{P}_i$, $\mathcal{N}_i$ denote the set of positive and negative indices for the $i$-th sample. The loss is then decomposed as follows.

\begin{align}
    \mathcal{L}_{repr} &= \frac{1}{\mid \mathbf{C}_s \mid} \biggl( \log_2 {\norm{\mathbf{C}_s}^2} - \log_2 {\norm{\mathbf{C}_{s} \odot \mathbf{C}_{y}}^2} \biggl) \\
    &= \frac{1}{\mid \mathbf{\mathbf{C}_s} \mid} \log_2 \sum_{i} \left( \sum_{j \in \mathcal{P}_i} \left( \mathbf{C}_{s} \right)_{ij}^2 + \sum_{j \in \mathcal{N}_i} \left( \mathbf{C}_{s} \right)_{ij}^2\right) \\ 
    &\;\;\;\; - \frac{1}{\mid \mathbf{\mathbf{C}_s} \mid} \log_2 \sum_{i} \sum_{j \in \mathcal{P}_i} \left( \mathbf{C}_{s} \right)_{ij}^2 \nonumber
\end{align}

This loss can be further simplified by the log identity $log(a) - log(b) = log(a/b)$.

\begin{align}
    \mathcal{L}_{repr} &= \frac{1}{\mid \mathbf{C}_s \mid} \log_2 \frac{\sum_{i} \left( \sum_{j \in \mathcal{P}_i} \left( \mathbf{C}_{s} \right)_{ij}^2 + \sum_{j \in \mathcal{N}_i} \left( \mathbf{C}_{s} \right)_{ij}^2\right)}{\sum_{i} \sum_{j \in \mathcal{P}_i} \left( \mathbf{C}_{s} \right)_{ij}^2} \\
    &= -\frac{1}{\mid \mathbf{C}_s \mid} \log_2 \frac{\sum_{i} \sum_{j \in \mathcal{P}_i} \left( \mathbf{C}_{s} \right)_{ij}^2}{\sum_{i} \left( \sum_{j \in \mathcal{P}_i} \left( \mathbf{C}_{s} \right)_{ij}^2 + \sum_{j \in \mathcal{N}_i} \left( \mathbf{C}_{s} \right)_{ij}^2\right)} \\
    % &= -\frac{1}{\mid \mathbf{K} \mid} \log_2 \sum_i \frac{\sum_{j \in \mathcal{P}_i} sim(z_i, z_j) }{\sum_k sim(z_i, z_j)}
    &= -\frac{1}{\mid \mathbf{C}_s \mid} \log_2 \sum_i \frac{\sum_{j \in \mathcal{P}_i} \left( \mathbf{C}_{s} \right)_{ij}^2}{\sum_k \left(\mathbf{C}_{s}\right)^2_{ik}}
\end{align}

The original supervised contrastive loss~\cite{Khosla2020SupervisedLearning} is given as follows.

\begin{align}
    \mathcal{L}_{SupCon} &= -\sum_i \log \frac{1}{\mid \mathcal{P}_i \mid} \frac{\sum_{j \in \mathcal{P}_i} sim(z_i, z_j) }{\sum_k sim(z_i, z_j)}
\end{align}

where $i,k \in \{1\dots |\mathbf{C}_s|\}$ index the set of all sampled pixels. In the case where we define $sim(z_i, z_j)$ to be the cosine similarity between the two vectors $z_i$ and $z_j$, these two losses are very similar. The only distinction between the two lies in switching the position of the normalisation and summation with respect to the logarithm. It is also worth noting that we use base 2 for the logarithm, as is convention in the information theory literature. In essence, the numerator in this loss pushes positive terms together, while the denominator repels negative pairs.

\newpage

% Rebuttal
\subsection{Comparison with other distillation methods} 
\begin{figure}[h]
\centering
\includegraphics[width=.95\linewidth]{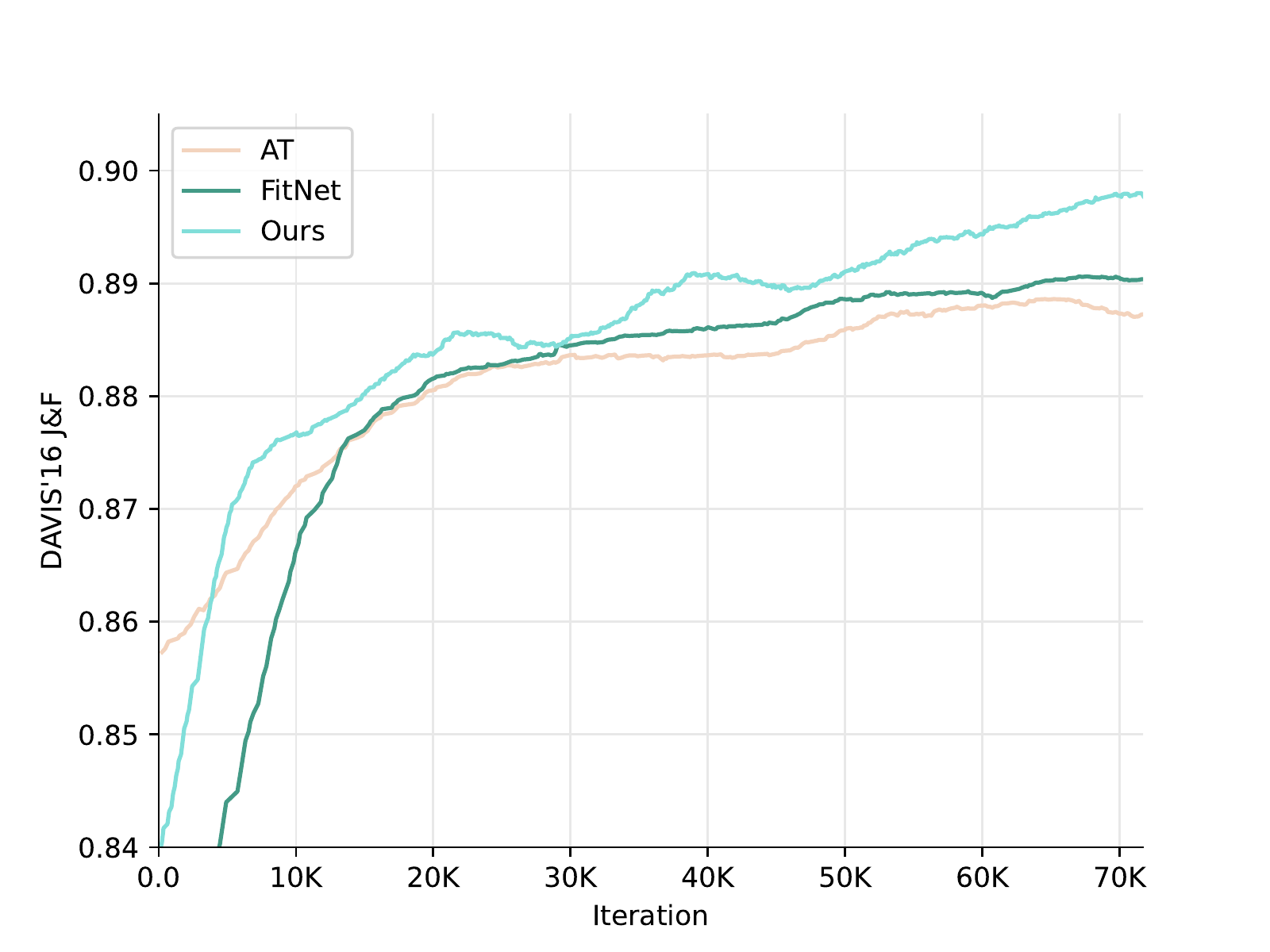}
\caption{Comparing the performance of the ResNet model trained with hints (FitNets)~\cite{Romero2015FitNets:Nets} and Attention Transfer~\cite{Zagoruyko2019PayingTransfer}.}
\label{fig:kd_comparison}
\end{figure}

We trained the ResNet model with two different distillation losses and observed a significant drop in attainable performance on the DAVIS16 benchmark, which can be seen in figure \ref{fig:kd_comparison}. Our method outperforms others by over 1 $\mathcal{J}\&\mathcal{F}$.

%We refer to table 4 (main text) for parameter count and runtime values on the same hardware, which compares our method with other methods.
\subsection{Loss ablation with the MobileNet backbone}
Additional experiments demonstrating the effectiveness of our proposed loss on the MobileNet architectures  are given in table \ref{tab:rebuttal}, whereby we observe a consistent improvement in $\mathcal{J}\&\mathcal{F}$ across both the DAVIS16 and DAVIS17 datasets with and without ASPP. 

                   % DAVIS 16                       DAVIS 17
% w/o exp_jk_11    89.74 J&F       89.11 J&F       79.86           79.11  
% ASPP exp_jk_12   89.60 J&F       88.0 J&F        81.84           80.30   

% w/0 exp_vag_11   89.18 J&F,      87.9 J&F,       80.48 J&F       78.93 J&F 
% ASPP exp_vag_13  89.6 J&F,       87.9 J&F,       81.63 J&F       79.97 J&F   

% w/o exp_a12      89.58 J&F       89.41 J&F       81.62 J&F       81.19 J&F    
% ASPP exp_a13     89.7 J&F        89.6 J&F       81.7 J&F        80.4 J&F         

\begin{table}[h]
\centering
\begin{tabular}{lccc}
    \toprule
    \textbf{Model} & distillation & DAVIS16 & DAVIS17 \\
    \midrule
    wo/ ASPP & \xmark & 89.2 & 80.5 \\
    wo/ ASPP & \cmark & \bf{90.1} & \bf{81.8} \\
    w/ ASPP & \xmark & 89.6 & 81.6 \\
    w/ ASPP & \cmark & \bf{90.5} & \bf{82.2} \\
    \bottomrule
\end{tabular}
\caption{Evaluating the effectiveness of our proposed distillation loss on the MobileNet backbone architecture.}
\label{tab:rebuttal}
\end{table}

%%%%%%%%% REFERENCES
% {\small
% \bibliographystyle{ieee_fullname}
% \bibliography{references}
% }

\end{document}